%% file: main.tex
\documentclass[10pt]{article} 

\PassOptionsToPackage{numbers, compress}{natbib}

\usepackage[preprint]{tmlr}

\input{math_commands}

\usepackage[utf8]{inputenc} 
\usepackage[T1]{fontenc}    
\usepackage{hyperref}       
\usepackage{url}            
\usepackage{booktabs}       
\usepackage{amsfonts}       
\usepackage{nicefrac}       
\usepackage{microtype}      
\usepackage{xcolor}         

\usepackage{tabularx}
\usepackage{multirow}
\usepackage{graphicx}
\usepackage{float}
\usepackage{wrapfig}
\usepackage{float}
\usepackage{amsmath}

\usepackage{amssymb}
\usepackage{bm}
\usepackage{wrapfig}
\usepackage{enumitem}
\usepackage{svg}
\usepackage{xspace}
\usepackage{pifont}
\usepackage{threeparttable}

\newcommand{\anja}[1]{{\color{blue}\textbf{[AR:} #1]}}

\newcommand{\keanu}[1]{{\color{magenta}\textbf{[KN:} #1]}}
\newcommand{\yes}{{\color{olive}\ding{51}}}%
\newcommand{\no}{{\color{red}\ding{55}}}%
\renewcommand{\anja}[1]{}
\renewcommand{\keanu}[1]{}
\makeatletter
\DeclareRobustCommand\onedot{\futurelet\@let@token\@onedot}
\def\@onedot{\ifx\@let@token.\else.\null\fi\xspace}

\def\eg{\emph{e.g}\onedot} 
\def\ie{\emph{i.e}\onedot} 
 
 \def\vs{\emph{vs}\onedot}

\newcommand{\DataShort}{AUDITS}
\newcommand{\DataLong}{Analysis Under Domain-shifts, qualIty, Type, and Size}
\newcommand{\DataLongU}{\underline{A}nalysis \underline{U}nder \underline{D}omain-shifts, qual\underline{I}ty, \underline{T}ype, and \underline{S}ize}


\title{Multi-axis Analysis of Image Manipulation Localization}

\author{
\name Keanu Nichols \email kmn5409@bu.edu \\
\addr Boston University
\AND
\name Divya Appapogu \email divsp@bu.edu \\
\addr Boston University
\AND
\name Giscard Biamby \email gbiamby@berkeley.edu \\
\addr University of California, Berkeley
\AND
\name Dina Bashkirova\thanks{Work done while at Boston University; currently at Google.} \email dbash@google.com \\
\addr Boston University
\AND
\name Anna Rohrbach \email anna.rohrbach@tu-darmstadt.de \\
\addr Technical University of Darmstadt \& hessian.AI
\AND
\name Bryan A. Plummer \email bplum@bu.edu \\
\addr Boston University
}

\begin{document}

\maketitle

\input{sec/0_abstract} 
\input{sec/1_intro}

\input{sec/2_previous_work}

\input{sec/3_method}

\input{sec/4_experiments}
\input{sec/5_conclusion}

\clearpage
\bibliography{main}
\bibliographystyle{tmlr}

\clearpage
\input{x_supplementary}




\end{document}

%% file: math_commands.tex
\usepackage{amsmath,amsfonts,bm}

\def\eqref#1{equation~\ref{#1}}

\def\1{\bm{1}}

\def\vs{{\bm{s}}}

\DeclareMathAlphabet{\mathsfit}{\encodingdefault}{\sfdefault}{m}{sl}
\SetMathAlphabet{\mathsfit}{bold}{\encodingdefault}{\sfdefault}{bx}{n}



%% file: sec/0_abstract.tex
\begin{abstract}

Advanced image editing software enables easy creation of highly convincing image manipulations, which has been made even more accessible in recent years due to advances in generative AI. Manipulated images, while often harmless, could spread misinformation, create false narratives, and influence people’s opinions on
important issues. Despite this growing threat, there is limited research on detecting advanced manipulations across different visual domains. Thus, we introduce \DataLongU{} (\DataShort{}), a comprehensive benchmark designed for studying axes of analysis in image manipulation detection. \DataShort{} comprises over 530K images from \emph{two distinct sources} (user and news photos). We curate our dataset to support analysis across multiple axes using recent diffusion-based inpaintings, spanning a diverse range of manipulation types and sizes. We conduct experiments under different types of \emph{domain shift} to evaluate robustness of existing image manipulation detection methods. Our goal is to drive further research in this area by offering new insights that would help develop more reliable and generalizable image manipulation detection methods
\footnote{\DataShort{} is publicly available at: \url{https://huggingface.co/datasets/DivyaApp/AUDITS}}.
\end{abstract}

%% file: sec/1_intro.tex
\section{Introduction}

\begin{figure}[t] 
  \centering
  \includegraphics[width=\textwidth,trim=2.3cm 4.5cm 1.5cm 4.5cm,clip]{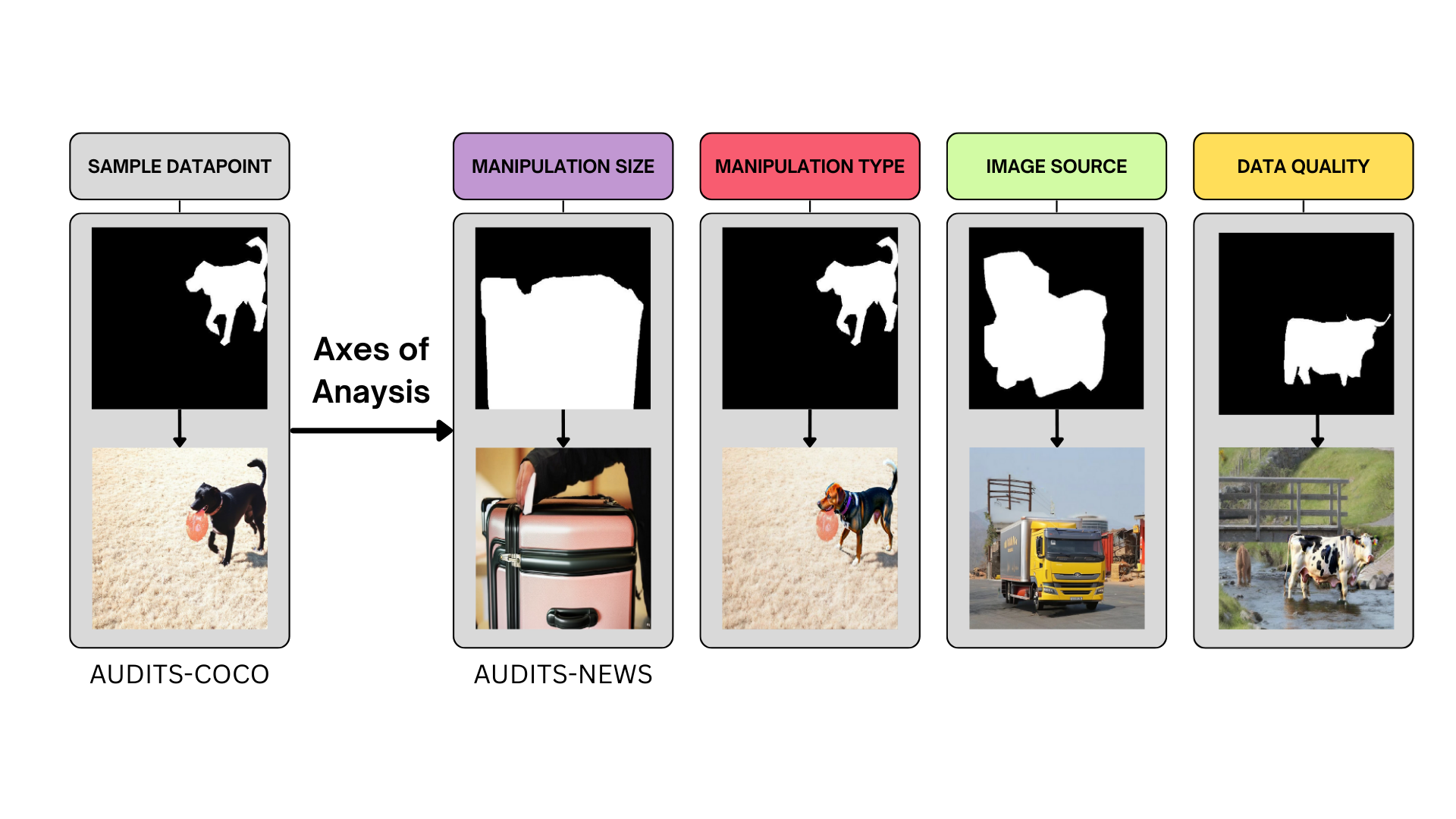} 
  \vspace{-6mm}
  \caption{
  We illustrate the axes of analysis within \DataShort{}, a comprehensive benchmark for image manipulation detection. Typically, manipulation datasets focus on one axis of analysis, namely unseen manipulation types (\eg Splicing \vs CopyMove). In contrast, \DataShort{} has four such axes: image source (COCO \vs VisualNews), data quality (\eg human perception label as low quality \vs high quality), manipulation type (\eg Blended Diffusion \vs Stable Diffusion) and manipulation size (\eg Small \vs Large), and is easily extended to other axes of analysis.}
  \label{fig:motivation_figure}
    \vspace{-4mm}
\end{figure}

Advanced image editing tools provide an ever widening access to realistic and easy to use mechanisms for creating convincing image edits with ease~\citep{genai1,genai3,genai5,dalle3, imagegen, imgman4}.  While this can be used for benign uses like satire and humor, these image edits can also help create false narratives like misrepresenting key individuals~\citep{Hifi}, resulting in the spread misinformation. To defend against this, many methods have been developed to detect manipulated images and/or localize what has been altered~\citep{PSCCNet, ObjectFormer,Hifi,mmFusion,evp,sida}.  While they have found that distribution shifts that stem from changing the type of manipulation used to alter the image or the size of the manipulated region, Figure~\ref{fig:motivation_figure} shows this only represents a subset of the challenges these models may face in practice.  For example, a model trained on a single image domain may incorrectly learn that any distribution shift is the result of a manipulation, resulting in many false positives on images from a new domain at test time.  However, as summarized in Table~\ref{tab:image_datasets}, existing benchmarks do not provide the ability to control for the many factors that may arise in applications of image manipulation localization (\eg,~\citet{Columbia, CASIAV1V2, COVERAGE, DEFACTO, dolos, CocoGlide-TruFor, TGIF,coco_inpaint}), making it impossible to accurately diagnose issues with current methods.

To address this shortcoming, we introduce \DataLongU{} (\textbf{\DataShort{}}), a carefully curated, large-scale image manipulation dataset.  The key contribution of this benchmark is the ability to evaluate the ability of methods to generalize across a wider range of settings like domain or image quality shifts that were not possible using prior work.  Specifically, \DataShort{} dataset is designed to enable exploration along four major axes: 1) measuring the effect of domain shifts on image manipulation detectors, 2) explore how manipulation quality (measured via a human perceptual study) impacts detection performance, 3) provide diverse manipulation types to ensure models can generalize across techniques, and 4) evaluate how the size of a manipulation impacts performance. These benefits enable us to study the effect on current methods with some intriguing findings.  For example, while many methods in prior work study the effect of generalizing across manipulation techniques (\eg~\citet{dolos,Mantranet,Span}), we find that a change in image sources also makes a significant impact to performance.  This is likely due, in part, to the localization models learning to correlate distribution shifts with manipulations, thereby incorrectly using shifts in image sources as supposed evidence of editing.  In addition, we find many methods are biased towards predicting images as not manipulated, resulting in models that struggle with detecting larger manipulations.  

One of the most striking observations from our experiments is that many of these issues stemming from our multi-axis shifts are not easily fixed with existing methods.  For example, domain generalization methods should make a model more robust shifts in image domain (\eg,~\citet{swad,modelsoups,miro,urm}), but we find they do not improve performance on \DataShort{}.  
This failure is likely attributed, in part, to the observed brittleness of these methods to multi-axis shifts in prior work~\citep{Humblot-Renaux_2024_CVPR,wangNAG2026}.  Further, while poor quality manipulations may seem like they should be easy for a model to detect, due to the ease as which human users could identify them, we find most models have negligible differences in detection performance due to the caliber of edit.  These issues highlight critical failures of current methods that would limit their use.  For example, a social media moderator would likely consider a detector unreliable after seeing it miss many obvious image manipulations.  These issues highlight the need for our dataset to directly evaluate and explore the ability of image manipulation detectors, while also providing insight into ways that they could be improved.

\begin{table*}[t]
    
    \centering
    \setlength{\tabcolsep}{2pt}
    \caption{We provide a comparison of our dataset, \DataShort{}, to several prior datasets w.r.t.: overall dataset size (Images), contains multiple domains , include diffusion manipulations, controls for sizes, number of manipulation types and includes a human perception label in their dataset. \DataShort{} stands out with its size, emphasis on diffusion-based inpainting, and inclusion of two image sources (COCO and News), surpassing the scope of previous datasets by incorporating four axes of analysis.}
    \label{tab:image_datasets}
    \begin{tabular}{lccc ccc}
    \toprule
    Dataset & Images & Multi  &    Diffusion   & Controlled & \# MT   &  w/Human  Perc-\\
            &         &      Domain        &  MTs & Sizes      &           & eption Label\\
    \midrule
    COLUMBIA~\citep{Columbia} & 1,845 & \no & \no & \no & 1 & \no  \\
    CASIAV1~\citep{CASIAV1V2}  & 1,721 & \no & \no  & \yes & 2 & \yes \\
    CASIAV2~\citep{CASIAV1V2}  & 12,323 & \no & \no & \yes & 2 & \yes  \\
    COVERAGE~\citep{COVERAGE}  & 200 & \no & \no & \no  & 1 & \no \\
    DEFACTO~\citep{DEFACTO}  & 229,000 & \no & \no & \no & 4 & \no   \\
    IMD2020~\citep{imd2020}  & 2,424 & \no  & \no & \no & 1 & \no   \\
    DOLOS~\citep{dolos}  & 148,112 & \no & \yes & \no & 5 & \no   \\
    TGIF~\citep{TGIF}
    & 75,000 & \no & \yes &  \no & 3 & \no  \\
    COCOGlide~\citep{CocoGlide-TruFor}  & 512 & \no & \yes & \no & 1 & \no \\
    AutoSplice~\citep{autosplicing}  & 5894 & \no & \yes & \no & 1 & \no \\
    MagicBrush~\citep{MagicBrush} & 10,388 & \no & \yes & \no & 1 & \yes \\
    COCO-Inpaint~\citep{coco_inpaint} & 375,000 & \no & \yes & \no & 6 & \no \\

    \midrule
    \DataShort{} (ours) & 530,640 & \yes & \yes & \yes & 11 & \yes \\
    \bottomrule
    \end{tabular}
\end{table*}

\begin{figure*}[t] 
  \centering
    \includegraphics[width=0.87\textwidth]{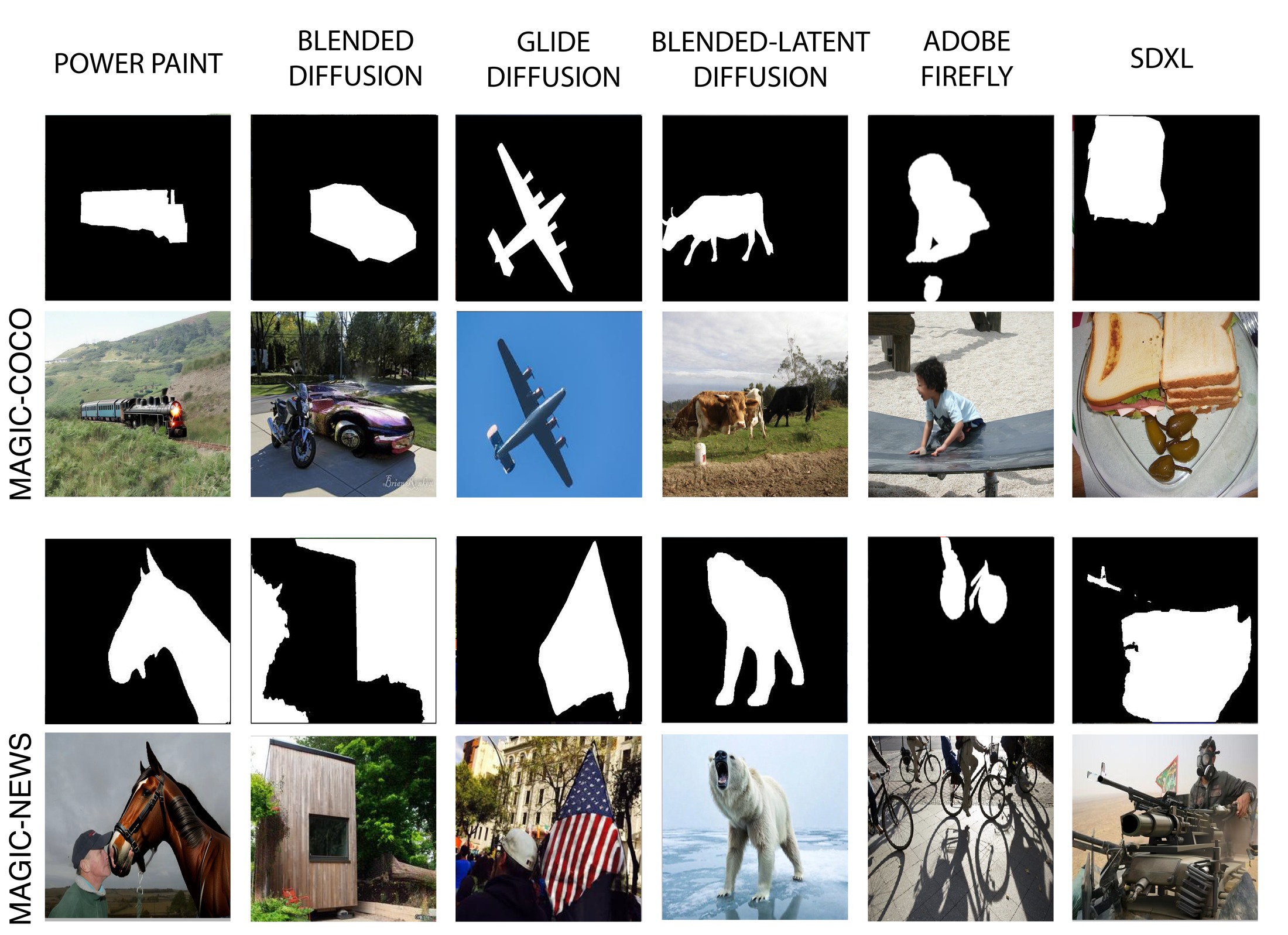}
  
    \vspace{-2mm}
  \caption{  
  Examples from our benchmark \DataShort{}, which contains 530,640 images from two visually distinct domains: COCO~\citep{mscoco} and Visual News~\citep{visualnews}. The dataset contains 11 state-of-the-art manipulation techniques classified into three categories: removal, replacement, and insertion. Manipulations cover a wide range of sizes, from 1\% to 100\% of the image area. These examples demonstrate the diversity of manipulation types and sizes in our dataset.
  }
  \vspace{-4mm}
  \label{fig:dataexamples}
\end{figure*}

In summary, our work makes the following contributions:
\vspace{-2mm}
\begin{itemize}[nosep,leftmargin=*]
    \item We introduce \DataShort{}, a large-scale image manipulation detection dataset consisting of 530,640 image-mask pairs. Our dataset has diverse distributions across four axes and is conducting using 11 image manipulation techniques.
    \item  Our analysis of existing methods shed light on such unique challenges posed by our dataset, such as a significant challenge in generalizing to OOD images, that could provide valuable insights for future research directions. 
    \item We explore potential solutions to some of our dataset's challenges, \eg integrating domain generalization techniques, and demonstrate that these solutions still struggle to boost performance, further highlighting a need for future work.
\end{itemize}

%% file: sec/2_previous_work.tex
\section{Related Work}

Table~\ref{tab:image_datasets} compares our \DataShort{} benchmark with datasets proposed in prior work including five main factors that indicate how well a dataset may reflect real-world applications.  Notably, prior work only ever covers a subset of the potential axes of analysis, limiting its usefulness. CASIA~\citep{CASIAV1V2} includes a human perception study and MagicBrush~\citep{MagicBrush} 
provides human annotations, but neither evaluates manipulation localization performance conditioned on human perception labels. In contrast, \DataShort{} includes a subset of images annotated by human-perceived manipulation quality, enabling analysis of performance 
with respect to perceptual realism. Another important factor is image source diversity. Having multiple source domains allows us to isolate the effect of domain shifts on image manipulation localization models.  However, many existing datasets are derived from a single domain such as faces or generic user images. For example, DOLOS~\citep{dolos} uses face datasets FFHQ~\citep{FFHQ} and CelebA~\citep{celebA}, whereas TGIF~\citep{TGIF} and CocoGlide~\citep{CocoGlide-TruFor} use COCO~\cite{mscoco}. COVERAGE~\citep{COVERAGE} consists primarily of consumer photography and therefore represents a single-domain dataset. However, using image editing to perpetuate misinformation via traditional journalism is a significant danger of these methods.  Thus, in our benchmark we have samples from both COCO and images in news articles~\citep{visualnews}, providing greater diversity than prior work that reflects real-world uses of these models.

We evaluate a range of image manipulation localization models with our benchmark that represents diverse design approaches~\citep{PSCCNet,dolos,Hifi,mmFusion,CocoGlide-TruFor,sida,swin,evp}.  Many of these methods evaluated their ability to generalize across Manipulation Types (MT) or editing sizes (\eg,~\citet{dolos,Hifi,sida}), but limitations in the benchmarks meant they could not accurately isolate the contribution of other factors like the quality of the manipulation or the effect of image sources.  In fact, sometimes models used external datasets that change both image sources and MTs to demonstrate an ability to generalize (\eg,~\citet{PSCCNet,mmFusion,evp}), but this coupling meant researchers would not understand if any drop in performance was due to MTs, image sources, or both.  Further, while the CASIA~\citep{CASIAV1V2} and MagicBrush~\citep{MagicBrush} datasets have human perception editing labels, they were only used by the original to validate the dataset's quality rather than providing insight into model performance like our work.   In contrast, we use these labels along with other controlled factors to provide more extensive and precise insights into model performance to better inform future work.

%% file: sec/3_method.tex
\section{\DataLong{} (\DataShort{})}
\label{sec:dataset}

Given an input image \( I \in \mathbb{R}^{H \times W \times 3} \), the goal of image manipulation localization is to learn a function \( g: I \rightarrow \hat{M} \) that predicts a manipulation mask \( \hat{M} \in [0,1]^{H \times W} \), estimating the ground-truth mask: 
    \[
    M_{i,j} =
    \begin{cases} 
    1, & \text{if pixel } (i,j) \text{ is manipulated} \\
    0, & \text{if pixel } (i,j) \text{ is authentic}
    \end{cases}
    \]
To better understand how performance on this task varies under different domains, we introduce our dataset \DataShort{} that comprises of two distinct subsets. \DataShort{}-News contains 311,989 images, with 271,527 manipulated images and 40,462 pristine images sampled from VisualNews~\citep{visualnews}. \DataShort{}-COCO contains 218,651 images, with 187,982 manipulated images and 30,669 pristine images sampled from COCO~\citep{mscoco}. All the manipulated images are accompanied by ground-truth masks. \anja{why do we care about this?}\keanu{Addressed}We design our benchmark such that we can study how different axes of analysis affect image manipulation localization performance across domains. 

\subsection{Image Manipulation Techniques}
\label{sec:manipulation_types}
\label{sec:maniuplation}
While constructing our dataset, we sought to generate a wide variety of image manipulations that would reflect real-world scenarios, offering a significant challenge for models tasked with localizing these alterations. To achieve this, we employed eleven major diffusion-based manipulation techniques: nine perform replacement, one performs removal, and one performs insertion. These methods are applied to both \DataShort{}-News and \DataShort{}-COCO, ensuring high diversity of manipulated images.  Additional details follow. 

\subsubsection{Replacement and Removal Based Manipulations}
\label{sec:replace_removal_manips}

We employ a diverse set of diffusion-based inpainting models to generate replacement and removal manipulations. These models span multiple generations of diffusion editing pipelines and exhibit different editing behaviors, enabling us to evaluate detector robustness across a wide range of manipulation techniques.

Our dataset includes several widely used text-guided diffusion models, including \textbf{Blended Diffusion}~\citep{blended_diffusion}, \textbf{Blended Latent Diffusion}~\citep{blended_latent_diff}, \textbf{Stable Diffusion}~\citep{stable_diffusion}, \textbf{SDXL}~\citep{SDXL}, and \textbf{GLIDE}~\citep{GLIDE}. These models support localized image editing through text-conditioned diffusion and latent-space inpainting. We also incorporate more recent diffusion-based editing frameworks such as \textbf{PowerPaint}~\citep{PowerPaint}, \textbf{Flux-Inpainting}~\citep{fluxinpainting}, and \textbf{HD-Painter}~\citep{hdpainter}, which introduce improved prompt conditioning, structural consistency, and harmonization mechanisms for producing visually coherent edits. In addition to replacement edits, we include removal-based manipulations using \textbf{Latent Diffusion}~\citep{stable_diffusion}, where the selected object region is removed and inpainted to blend with the surrounding scene. Finally, we include \textbf{Adobe Firefly}~\citep{adobe_firefly}, a proprietary diffusion-based editing system, to better reflect real-world image editing tools that are widely accessible to users.

%
All these models are applied to both the \DataShort{}-News and \DataShort{}-COCO images. For \DataShort{}-News, segmentation masks are generated using ODISE~\citep{ODISE}, whereas for \DataShort{}-COCO, the existing masks provided by the COCO dataset are used. 
All the models use both text and image data to guide the generation of manipulated images. We identify an object randomly selected from the panoptic mask and replace that region with the new content based on the object class using the object's mask. For example, if the selected mask corresponds to a ``cat'', the information that the object is a cat is used as a prompt to guide the diffusion models in generating the new content (\ie, another cat).

\subsubsection{Insertion-Based Manipulations}

We also insert/splice new objects into an existing images. We start splicing by utilizing ODISE to generate panoptic segmentation masks from images sourced from VisualNews and for MS~COCO we use the given panoptic masks. Panoptic segmentation provides a detailed and comprehensive breakdown of the scene by categorizing each pixel into specific objects (things) and background elements (stuff). Once the segmentation map is obtained, \textbf{GLIGEN}~\citep{gligen} is employed to generate a new object that fits contextually into the designated area. GLIGEN leverages the spatial and semantic information from the segmentation map to ensure that the newly generated object not only aligns with the surrounding elements in terms of position and scale but also blends seamlessly with the scene's overall aesthetics. The generated object is then carefully inserted into the original image \( I_{\text{original}} \) using the object mask \( M \) (Here the object mask  \( M \) is randomly chosen from the panoptic segmentation), resulting in a spliced image \( I_{\text{spliced}} \), as described by:
\(I_{\text{spliced}} = \text{GLIGEN}(S) \odot M \oplus I_{\text{original}}\). Where \( S \) is the segmentation map, \( M \) is the object mask, and \( I_{\text{original}} \) is the original image.

\subsection{Diversity in \DataShort{} }
\label{sec:topic}

When constructing the \DataShort{}-dataset, we curate a wide range of scenarios to better capture the different challenges that an image manipulation detector may encounter in real-world applications. News imagery, in particular, covers an broad variety of content, as shown in Figure~\ref{fig:dataexamples}. To ensure diversity in image content, we select our images from 13 different topics from the VisualNews dataset (further details in Section~\ref{sec:audits_news_stats} of the Appendix). The VisualNews dataset draws from four major news outlets: USA Today, Washington Post, BBC, and The Guardian. We sample a substantial number of images from each topic and news outlet, details of this can be found in Section~\ref{sec:audits_news_stats} of our Appendix.
MS~COCO does not have topic labels; for our \DataShort{}-COCO subset we sample 82 object categories, including the most common objects (\textit{person}, \textit{car}), and less frequent objects (\textit{hairbrush}, \textit{giraffe}). The combination of news-related imagery and everyday objects ensures that our dataset represents not just specialized journalistic content but also a broad spectrum of general, real-world scenes.

In the \DataShort{}-News we manipulate both the stuff (\eg sky, terrain) and thing (\eg person, cat) categories, resulting in a wide variety of manipulation sizes based on the panoptic segmentation masks predicted by ODISE. This is in contrast to prior work that often focuses solely on foreground object manipulation~\citep{DEFACTO, imd2020}.

Our \DataShort{}-dataset exhibits diversity in manipulation sizes. 
The \DataShort{}-News subset shows a wide range of manipulation sizes, from small edits to large alterations, requiring detection models to be robust across various levels of visual modifications. In contrast, manipulation sizes in \DataShort{}-COCO are skewed toward smaller edits, with relatively few medium and large manipulations. 
In \DataShort{}-News, there are 142,770 small manipulations covering < 25\%  of the image area, 107,872 medium manipulations covering between 25\% and 60\% of the image area, and 20,885 large manipulations covering > 60\% of the image area. For \DataShort{}-COCO, the breakdown is 155,106 small, 27,553 medium, and 5,723 large manipulations, respectively. More details are in Section~\ref{sec:mask_size_distribution} of the Appendix.

\subsection{Dataset Quality Survey}
\label{sec:dataset_quality_survey}
To assess the quality of the manipulated images and ensure their practical use for manipulation detection, we perform a human evaluation via Amazon Mechanical Turk. A total of 4,950 images were used in our survey (2,750 from \DataShort{}-News and 2,200 from \DataShort{}-COCO). We include 5 generators for \DataShort{}-News and 4 for \DataShort{}-COCO\footnote{This evaluation included all manipulation types with the exception of Adobe Firefly (for either subset) and Blended Latent Diffusion / GLIGEN (for \DataShort{}-COCO).}. We sample 500 images from each generator. Additionally we include 250 authentic images from \DataShort{}-News and 200 from \DataShort{}-COCO. Each image is evaluated by 3 people; in total we have 14,850 responses from 1,829 unique workers. 
The respondents were shown two images: a manipulated image $\bm{X}$ modified in a region specified by a binary mask $\bm{M}$, and a copy of  $\bm{X}$ with a mask superimposed on it with a transparency value $\alpha$.  Additional details and examples are in Section~\ref{sec:human_results_section} of the Appendix.

\subsubsection{Survey Questions Details}
The respondents are asked the following questions: Q1) \textit{``Do you think this image is manipulated?''} (yes or no), Q2) \textit{``Do you see the \textbf{object} in the image (you can use the mask overlay to the right of the Image to better see the object)?''} (yes or no); Q3) \textit{``Does the \textbf{object} look realistic?''} (yes, maybe or no), and Q4) \textit{``Does the \textbf{object} look natural in the background?''} (yes, maybe, no). For each example, we specify which \textbf{object} the respondents should look for (\eg a ``giraffe''). We discard the answers to Q3 and Q4 if the respondent answered ``No'' to Q2, which results in 9,596 responses. Next, we utilize Q3 to create pseudo-labels for ``High quality'' and ``Low quality'' images. By combining the ``Maybe'' and ``No'' responses and utilizing majority voting, we get 1,507 ``Low quality'' images. Meanwhile, we have 1,905 ``Yes'' responses corresponding to ``High quality'' images, hence we get 3,412 images in total. 
We evaluate the effect of these pseudo-labels on manipulation detection models in Section~\ref{sec:quality_eval}.

%% file: sec/4_experiments.tex
\section{Experiments}\label{sec:experiments}
 
\begin{figure*}[t]
    \centering
    \includegraphics[width=0.88\textwidth]{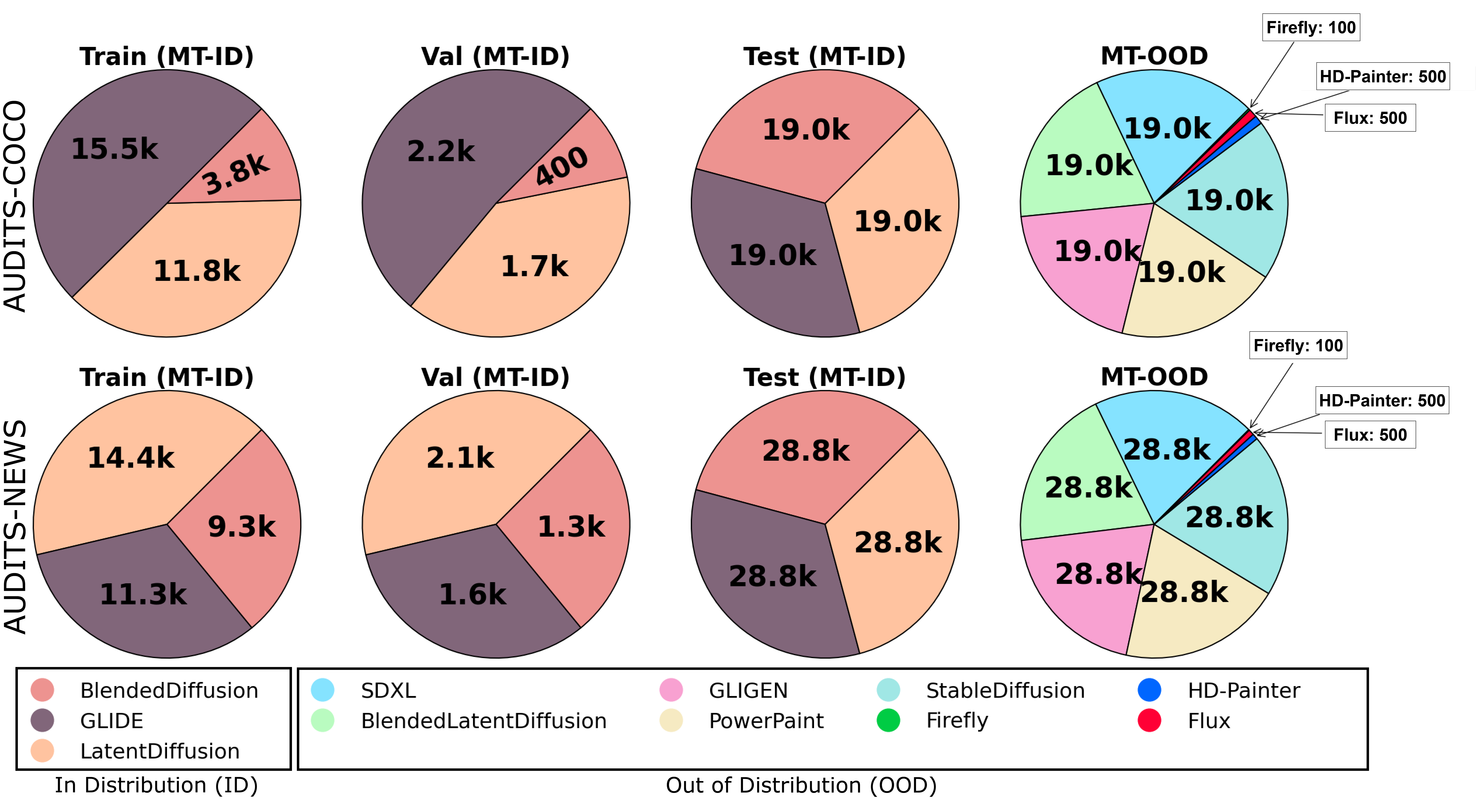} 
    \vspace{-4mm}
    \caption{Statistics on manipulation types in \DataShort{}-News and \DataShort{}-COCO.\anja{I found the IM-ID/MT-ID/etc confusing and not aligned with the figure here; is it even necessary to mention all of them?} \anja{some text is white on green - hard to see it} \anja{MAGIC-COCO, MAGIC-News, add ``-''}\keanu{Addressed}}
    \vspace{-4mm}
    \label{fig:data_breakdown_diagram}
\end{figure*}

To benchmark the recent manipulation detection models, we employ two different data splits based on manipulation type (11 types) and image subset (News or COCO): one for in-distribution (ID) and another for out-of-distribution (OOD) performance assessment. We compare a number of different image manipulation detection models in our experiments, namely: \textbf{PSCC-Net}~\citep{PSCCNet}, \textbf{MMFusion}~\citep{mmFusion}, \textbf{HiFi}~\citep{Hifi} ,\textbf{EVP}~\citep{evp}, \textbf{TruFor}~\citep{CocoGlide-TruFor}, \textbf{SIDA}~\citep{sida}, \textbf{DOLOS}~\citep{dolos}, \textbf{Swin Transformer}~\citep{swin} used as the base encoder
for an Upernet model \citep{upernet} and additionally utilize domain generalization methods like \textbf{SWAD}~\citep{swad}, \textbf{Model Soups}~\citep{modelsoups}, \textbf{MIRO}~\citep{miro} and \textbf{URM}~\citep{urm}. 
Figure~\ref{fig:data_breakdown_diagram} outlines the distribution of manipulation types across the \DataShort{}-News and \DataShort{}-COCO subsets, separating manipulation methods into those for training and those reserved for testing as out-of-domain\anja{neither is in the figure}. 

\subsection{Evaluation Protocol}
Manipulation localization is formulated as a pixel-level prediction task. Given an input image $I$, a detector produces a probability map $\hat{M} \in [0,1]^{H \times W}$ indicating the likelihood that each pixel belongs to a manipulated region. Since predictions are made at the pixel level, we evaluate models using \textbf{pixel-level micro-averaged metrics}, pooling predictions across all images in the dataset.

A straightforward evaluation would compare the predicted manipulation map $\hat{M}$ with the binary ground-truth mask $M$. However, diffusion-based editing models frequently introduce visual changes outside the annotated manipulation mask. These changes may arise from global adjustments such as color harmonization, texture blending, or lighting modifications that occur during the editing process. As a result, detectors may correctly identify regions that differ from the original image even though they fall outside the annotated mask.

Ignoring such regions would penalize detectors that identify real pixel-level changes, while treating them as fully manipulated could reward detections unrelated to the intended edit. To address this ambiguity, we introduce an intermediate category capturing \emph{off-mask drift}.

\paragraph{Ternary ground-truth representation.}
Let $O$ denote the original image and $E$ the edited image. We compute a per-pixel RGB difference map

\begin{equation}
D_{i,j} =
\frac{1}{3}\sum_{c \in \{R,G,B\}}
(O_{i,j,c} - E_{i,j,c})^2 .
\end{equation}

Using this difference map and the binary manipulation mask $M$, we construct a ternary ground-truth representation $G$:

\begin{equation}
G_{i,j} =
\begin{cases}
\text{manipulated}, & M_{i,j} = 1, \\
\text{ambiguous}, & M_{i,j} = 0 \ \text{and} \ D_{i,j} > \tau, \\
\text{authentic}, & \text{otherwise}.
\end{cases}
\end{equation}

The threshold $\tau$ determines whether visual differences outside the annotated manipulation mask are significant enough to be considered ambiguous. In practice, we compute the mean squared difference between the original and edited images and mark pixels with $D_{i,j} > \tau$ as ambiguous. In all experiments we set $\tau = 0.0025$, which was chosen empirically to detect visible editing artifacts while avoiding small numerical differences. The ambiguous category therefore captures pixels outside the annotated manipulation mask whose appearance differs from the original image. These differences arise from contextual modifications introduced by modern editing pipelines (\eg diffusion-based editing), which may alter surrounding pixels to ensure visual consistency. Identifying these pixels allows us to measure the effect of off-mask drift introduced during the editing process.

One possible way to eliminate ambiguous pixels would be to simply copy the manipulated region into the original image, ensuring that only pixels inside the annotated manipulation mask are modified. This has two major drawbacks.  First, this is not the expected usage by general users (\ie, most users would use the diffusion-based edit directly), thereby introducing a difference in evaluation \vs its practical use. Second, this procedure would artificially remove contextual changes
introduced by modern editing pipelines, particularly diffusion-based editing methods that often modify surrounding pixels to maintain visual consistency (\eg blending or color harmonization)  Thus, these edits are often easily identified by humans. In contrast, our goal is to evaluate detectors under realistic editing conditions rather than artificially constrained manipulations, and therefore we retain these off-mask modifications.

\paragraph{Binary evaluation with ambiguous-pixel weighting.}
Since common evaluation metrics such as AUROC, precision, recall, and F1 are defined for binary classification, we map the ternary labels to a binary prediction task. Pixels labeled as \emph{manipulated} are treated as positive examples, while pixels labeled as \emph{authentic} are treated as negative examples.

Ambiguous pixels fall outside the annotated manipulation mask and are therefore treated as negative examples. However, because these pixels may still reflect visual changes introduced by the editing process, we assign them reduced weight during evaluation. Specifically, we define a per-pixel weight function

\begin{equation}
w(i,j) =
\begin{cases}
1, & \text{if pixel } (i,j) \text{ is manipulated}, \\
\alpha, & \text{if pixel } (i,j) \text{ is ambiguous}, \\
1, & \text{if pixel } (i,j) \text{ is authentic},
\end{cases}
\end{equation}

where $\alpha \in [0,1]$ controls the contribution of ambiguous pixels. In our experiments we set $\alpha = 0.5$, which reduces the influence of ambiguous regions while still allowing them to contribute to the evaluation. This weighting scheme allows the evaluation to emphasize precise localization of the annotated manipulation region while reducing the impact of ambiguous off-mask changes. If the editing process modifies only pixels inside the annotated manipulation mask (\eg copy–paste manipulations), the ambiguous category is empty and the evaluation reduces to standard binary manipulation localization. These weights are incorporated into the Area Under the ROC Curve (AUROC), precision, recall, and F1 calculations described below.

\paragraph{Metrics.}
AUROC is computed using a streaming histogram estimator that approximates

\begin{equation}
\text{AUROC} = P(s_p > s_n) + \frac{1}{2}P(s_p = s_n),
\end{equation}

where $s_p$ and $s_n$ denote prediction scores for positive and negative pixels, respectively. The factor $1/2$ accounts for tied prediction scores,
as is standard in ROC analysis. Precision, recall, and F1 score are computed after thresholding prediction scores at $\mu = 0.5$. All metrics are computed using pixel-level micro averaging across the dataset.

The ambiguous-pixel weighted evaluation described above is used only for the pixel-level localization experiments in Section~\ref{sec:image_source_and_manip_type_gen} and Section~\ref{sec:manip_size_graphs}. For the rest of the tables we report standard AUROC and or F1 without ternary mask construction or ambiguous-pixel weighting. This keeps those results directly comparable to prior work while reserving the ambiguous-pixel weighted protocol for experiments where ambiguity outside the annotated manipulation mask is central to the analysis.

\subsection{Effect of Image Source Domain and Manipulation Type}
\label{sec:image_source_and_manip_type_gen}

\begin{table*}[t]
    \centering
      \setlength{\tabcolsep}{1.pt}
    \caption{Comparing AUC and F1 scores of models trained on \DataShort{}-News or \DataShort{}-COCO and evaluated on both datasets to measure generalization across image sources and manipulation types. Methods are grouped by base architecture to facilitate direct comparison with their domain generalization (DG) variants. Bold numbers indicate the best performance among baseline methods, while boxed values highlight the best result within each method family (e.g., EVP and its DG variants, MMFusion and its DG variants).}
    \begin{tabular}{l cccc cccc cccc cccc}
    \toprule
    Trained on: & \multicolumn{4}{c}{\DataShort{}-News} & \multicolumn{4}{c}{\DataShort{}-COCO} & \multicolumn{4}{c}{\DataShort{}-COCO} & \multicolumn{4}{c}{\DataShort{}-News}\\
    \midrule
    Tested on: & \multicolumn{8}{c}{\DataShort{}-News}  & \multicolumn{8}{c}{\DataShort{}-COCO}\\
    \midrule
     & \multicolumn{2}{c}{MT-ID} & \multicolumn{2}{c}{MT-OOD} & \multicolumn{2}{c}{MT-ID} & \multicolumn{2}{c}{MT-OOD} & \multicolumn{2}{c}{MT-ID} & \multicolumn{2}{c}{MT-OOD} & \multicolumn{2}{c}{MT-ID} & \multicolumn{2}{c}{MT-OOD}\\
     \midrule
     & AUC & F1 & AUC & F1 & AUC & F1 & AUC & F1 & AUC & F1 & AUC & F1 & AUC & F1 & AUC & F1\\
     \midrule
       
        \multicolumn{17}{l}{\textbf{EVP}} \\
        \midrule
        EVP  &93.2&75.6&\framebox[0.8cm]{\textbf{81.6}}&\framebox[0.8cm]{52.8}&74.4&37.9&\textbf{63.5}&20.8&85.9&53.2&\framebox[0.8cm]{76.7}&29.8&85.8&\textbf{57.4}&\framebox[0.8cm]{66.3}&\framebox[0.8cm]{27.3}\\
        EVP+SWAD              &86.1&62.0&77.3&45.4&77.6&41.9&63.1&22.3&86.8&56.0&72.1&\framebox[0.8cm]{30.1}&77.1&45.5&64.0&24.8\\
        
        EVP+Soup     &92.1&70.1&77.7&43.1&83.3&49.8&63.2&15.1&90.5&60.3&71.8&28.9&76.1&39.2&63.4&21.3\\
        
        EVP+MIRO              &\framebox[0.8cm]{95.4}&\framebox[0.8cm]{81.5}&76.4&44.0&81.7&\framebox[0.8cm]{55.9}&61.6&\framebox[0.8cm]{22.6}&\framebox[0.8cm]{94.4}&\framebox[0.8cm]{72.0}&73.2&25.8&\framebox[0.8cm]{92.1}&\framebox[0.8cm]{67.5}&64.5&22.8\\
        EVP+URM              &90.2&62.9&72.6&26.5&\framebox[0.8cm]{84.2}&50.9&\framebox[0.8cm]{63.5}&12.2&91.2&64.6&69.9&16.5&83.8&52.2&64.4&12.8\\
        \midrule
        \multicolumn{17}{l}{\textbf{MMFusion}} \\
        \midrule
        MMFusion &\framebox[0.8cm]{\textbf{96.8}}&\textbf{82.6}&73.8&51.5&\framebox[0.8cm]{\textbf{88.6}}&\textbf{70.9}&56.5&31.3&\textbf{96.7}&\textbf{75.0}&64.4&30.3&\framebox[0.8cm]{\textbf{92.3}}&\framebox[0.8cm]{56.7}&57.0&22.8\\
        MMFusion+SWAD    &96.6&82.5&\framebox[0.8cm]{80.7}&\framebox[0.8cm]{60.4}&86.8&\framebox[0.8cm]{71.4}&\framebox[0.8cm]{62.7}&\framebox[0.8cm]{38.9}&96.4&77.0&\framebox[0.8cm]{71.0}&\framebox[0.8cm]{38.9}&91.8&54.0&\framebox[0.8cm]{61.0}&\framebox[0.8cm]{28.7}\\
        MMFusion+MIRO         &96.5&\framebox[0.8cm]{82.7}&75.9&55.5&87.0&69.1&58.2&24.2&\framebox[0.8cm]{96.7}&\framebox[0.8cm]{78.4}&70.6&29.7&89.3&50.4&50.8&21.7\\
        \midrule
        \multicolumn{17}{l}{\textbf{Other Manipulation Detection Methods}} \\
        \midrule
        PSCC-Net  &89.1&70.5&58.9&33.2&70.7&51.1&45.7&40.6&80.2&45.7&68.3&33.6&85.2&49.7&64.0&25.3\\
        SIDA  &82.5&72.0&75.2&\textbf{62.4}&68.4&52.2&62.4&41.1&81.0&67.1&\textbf{76.8}&\textbf{60.0}&80.1&57.1&\textbf{67.9}&\textbf{41.9}\\
        TruFor  &59.4&36.1&55.7&37.9&49.8&44.1&48.0&\textbf{44.3}&49.7&25.9&47.7&25.9&64.2&30.5&56.7&25.6\\
        DOLOS &69.0&38.6&61.6&26.6&54.2&5.0&55.7&29.1&66.2&7.5&58.5&25.3&60.2&26.9&44.7&16.1\\
        HiFi &55.1&16.5&50.2&1.9&60.4&45.0&51.4&24.3&75.0&48.7&57.3&26.0&52.9&9.7&49.8&0.8\\
        Upernet &74.2&62.6&64.1&44.3&67.2&51.2&53.4&14.7&77.3&64.5&54.3&16.4&68.6&43.0&53.7&22.0\\
        
        \midrule
        
    \end{tabular}
      \vspace{-4mm}
    \label{tab:vn_news_results}
\end{table*}

Table~\ref{tab:vn_news_results} reports results when training models on \DataShort{}-News or \DataShort{}-COCO and evaluating on both datasets, allowing us to analyze generalization across manipulation types (MT) and image sources. Overall, performance degrades when models are evaluated on out-of-distribution settings arising from either MT or domain shifts. To isolate the effect of manipulation type, we compare performance when training and testing on the same image source while changing MT. For example, MMFusion trained and evaluated on \DataShort{}-News has a 23-point decrease. To isolate the effect of image source, we compare performance across datasets while keeping manipulation types fixed. When training on \DataShort{}-News and testing on \DataShort{}-COCO (MT-ID), MMFusion has an 8-point decrease. These results suggest that both manipulation type and image source shifts play an important role in building robust detectors. However, most prior work has primarily focused on generalization across MT~\citep{COVERAGE, DEFACTO, dolos, CocoGlide-TruFor, TGIF}, without isolating the effect of image source shifts.

We further compare each base model with its domain generalization (DG) variants to assess whether DG techniques improve robustness to these distribution shifts. For EVP, methods such as SWAD, MIRO, and URM yield comparable performance to the base model, with no single DG approach consistently outperforming EVP across all evaluation settings. A similar trend is observed for MMFusion, where DG variants 
provide occasional improvements but do not clearly dominate the base model.

Overall, these results suggest that standard DG methods do not consistently improve performance under the multi-axis distribution shifts considered in \DataShort{}. While MIRO~\cite{miro} and URM~\cite{urm} provide slight improvements for EVP in some settings, their gains are modest and inconsistent. Notably, URM is a more recent DG method, yet its improvements remain limited, suggesting that advances on standard DG benchmarks do not directly translate to improved performance on \DataShort{}.

\subsection{Effect of Manipulation Size}
\label{sec:manip_size_graphs}
Another aspect of our dataset is the manipulation size. Here, we determine how SoTA models perform across different sizes of manipulations. We utilize the full range of sizes for \DataShort{}-News which ranges from 20\% - 80\% and \DataShort{}-COCO which ranges from 0\% - 100\%.

\begin{figure*}[t] 
  \centering
  \includegraphics[width=0.9\linewidth]{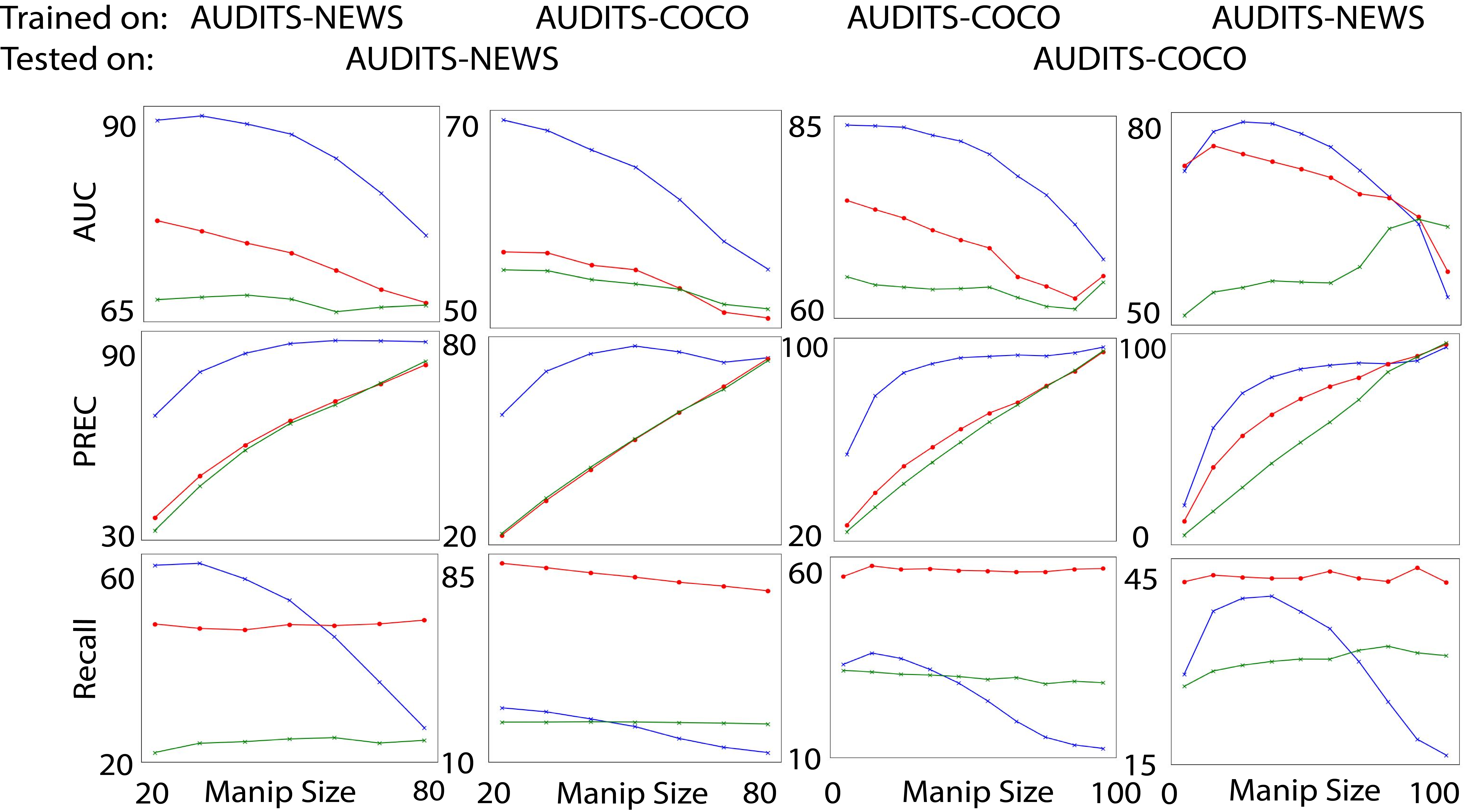} 
  \caption{Comparing the AUC/Precision/Recall performance of models trained on the \DataShort{}-News and \DataShort{}-COCO to study generalization across image distributions and manipulation sizes. \textcolor{blue}{EVP} results are in red, \textcolor{red}{PSCC-Net} are blue and \textcolor{green}{DOLOS} are green. The first two columns are results on \DataShort{}-News’s ID test set and OOD set by getting the average of all the results in the specific size category. The last 2 columns are tested on \DataShort{}-COCO’s ID test set and OOD set by getting the average of all the results in the specific size category\anja{same}\keanu{addressed}.\anja{make clear what is tested on -- in the table}\keanu{addressed}}
  \label{fig:ms_coco_size}
\end{figure*}

\noindent\textbf{Results.} Figure~\ref{fig:ms_coco_size} highlights the performance of a range of models, one somewhat counterintuitive observation is that AUC performance on small manipulations is generally higher than large ones.  However, this is simply due to the fact that the majority of the image is authentic, so any model that is biased towards predicting that image regions are real would get higher performance.  This also manifests itself in the very low recall rates of larger manipulations and corresponding high precision rates on these larger manipulations models are simply less likely to prediction an image region as manipulated unless it is fairly certain it is correct.

Comparing across methods, we see that EVP consistently gets better performance when evaluated on \DataShort{}-News.  However, we obtain mixed results on \DataShort{}-COCO when trained with \DataShort{}-News, where best performance bounces back and forth mostly between EVP and PSCC-Net.  This helps demonstrate the different properties of both splits, where the \DataShort{}-COCO manipulations contain a higher portion of small manipulations (see Section~\ref{sec:mask_size_distribution} of the Appendix for size statistics), which can result in significant degradation when training with the generally larger manipulations in \DataShort{}-News.

\subsection{Generalizing Across Inpainting Datasets}

We further evaluate whether models trained on \DataShort{} transfer to external diffusion-based inpainting datasets. Tables~\ref{tab:magicbrush} and ~\ref{tab:cocoglide} report results on MagicBrush and CocoGLIDE, respectively. These datasets differ from AUDITS in both image source and editing procedure, and therefore provide a complementary test of out-of-dataset generalization.

On MagicBrush, EVP substantially outperforms the other evaluated methods, improving AUC by more than 20 points over the next best method. However, the relative behavior of methods changes on CocoGLIDE: PSCC-Net performs competitively with the previously reported CocoGLIDE result, while DOLOS remains substantially lower. EVP is again the strongest method, outperforming PSCC-Net by roughly 6 AUC points.

These results show that method rankings are not fixed across external inpainting datasets. A method that performs poorly on one external dataset may perform competitively on another, suggesting that evaluation on a single dataset can give an incomplete picture of generalization. This further motivates \DataShort{} as a multi-axis benchmark for analyzing robustness across image sources, manipulation types, and editing conditions, rather than only reporting aggregate performance on one external test set.

\begin{table*}[t]
\centering

\begin{minipage}{0.48\textwidth}
\centering
\caption{Performance when evaluated on MagicBrush~\citep{MagicBrush} trained on \DataShort{}-News}
\label{tab:magicbrush}
\begin{tabular}{@{}lccc@{}}
\toprule
Model        & AUC  & F1   & IoU  \\ \midrule
PSCC-Net     & 46.7 & 25.6 & 11.8 \\
DOLOS        & 52.9 & 26.0 & 11.1 \\
EVP & \textbf{77.0} & \textbf{36.8} & \textbf{23.3} \\ 
\bottomrule
\end{tabular}

\end{minipage}
\hfill
\begin{minipage}{0.48\textwidth}
\centering
\setlength{\tabcolsep}{2pt}
\caption{Performance on CocoGLIDE~\citep{CocoGlide-TruFor} (trained on \DataShort-News).}
\label{tab:cocoglide}
\begin{tabular}{@{}lccc@{}}
\toprule
Model                           & AUC  & F1   & IoU  \\ \midrule
PSCC-Net (reported in CocoGLIDE) & 77.7 & 51.5 & -    \\
TruFor (proposed in CocoGLIDE)   & 75.2 & 52.3 & -    \\
PSCC-Net                        & 78.0 & 51.3 & 38.2 \\
DOLOS                           & 55.6 & 38.5 & 25.3 \\
EVP                             & \textbf{83.6} & \textbf{57.0} & \textbf{42.9} \\ 

\bottomrule
\end{tabular}
\end{minipage}

\end{table*}

\subsection{Effect of Manipulation Quality}
\label{sec:quality_eval}

\begin{table*}[t]

    \centering
    \caption{Comparing the AUC results from our human evaluation by utilizing majority voting we use 3412 images (1,507 ``Low'' quality and 1905 ``High'' quality images)  from our test and OOD set based on question 3 and combined answer choice Maybe with No. An image is considered of ``High'' quality if more people consider it realistic and ``Low'' if more people consider it not-realistic. The first 4 columns of results are tested on \DataShort{}-News's ID test set and OOD set combined and the last 4 columns of results are tested on \DataShort{}-COCO's ID test set and OOD set combined.}
    \begin{tabular}{lcccccccc}
    \toprule
    Trained on: & \multicolumn{2}{c}{\DataShort{}-News }& \multicolumn{2}{c}{\DataShort{}-COCO } & \multicolumn{2}{c}{\DataShort{}-COCO } & \multicolumn{2}{c}{\DataShort{}-News }\\
    \midrule
    Tested on: & \multicolumn{4}{c}{\DataShort{}-News }& \multicolumn{4}{c}{\DataShort{}-COCO }\\
    \midrule
    Model & Low & High & Low & High & Low & High & Low & High \\ 
    \midrule
        EVP      &80.9&80.7&58.9&59.5&90.3&88.0&69.3&68.0 \\
        DOLOS    &80.3&79.8&75.4&75.1&66.1&64.5&66.1&64.5 \\
        PSCC-Net &72.8&71.8&47.0&48.9&78.2&77.2&49.3&49.0 \\
        HiFi     &72.6&72.3&48.4&48.7&73.3&71.5&51.5&51.4 \\
       \bottomrule
    \end{tabular}
    \label{tab:survey_results_labelled_test}
\end{table*}

\begin{figure*}[t]
    \centering
    \includegraphics[width=\textwidth]{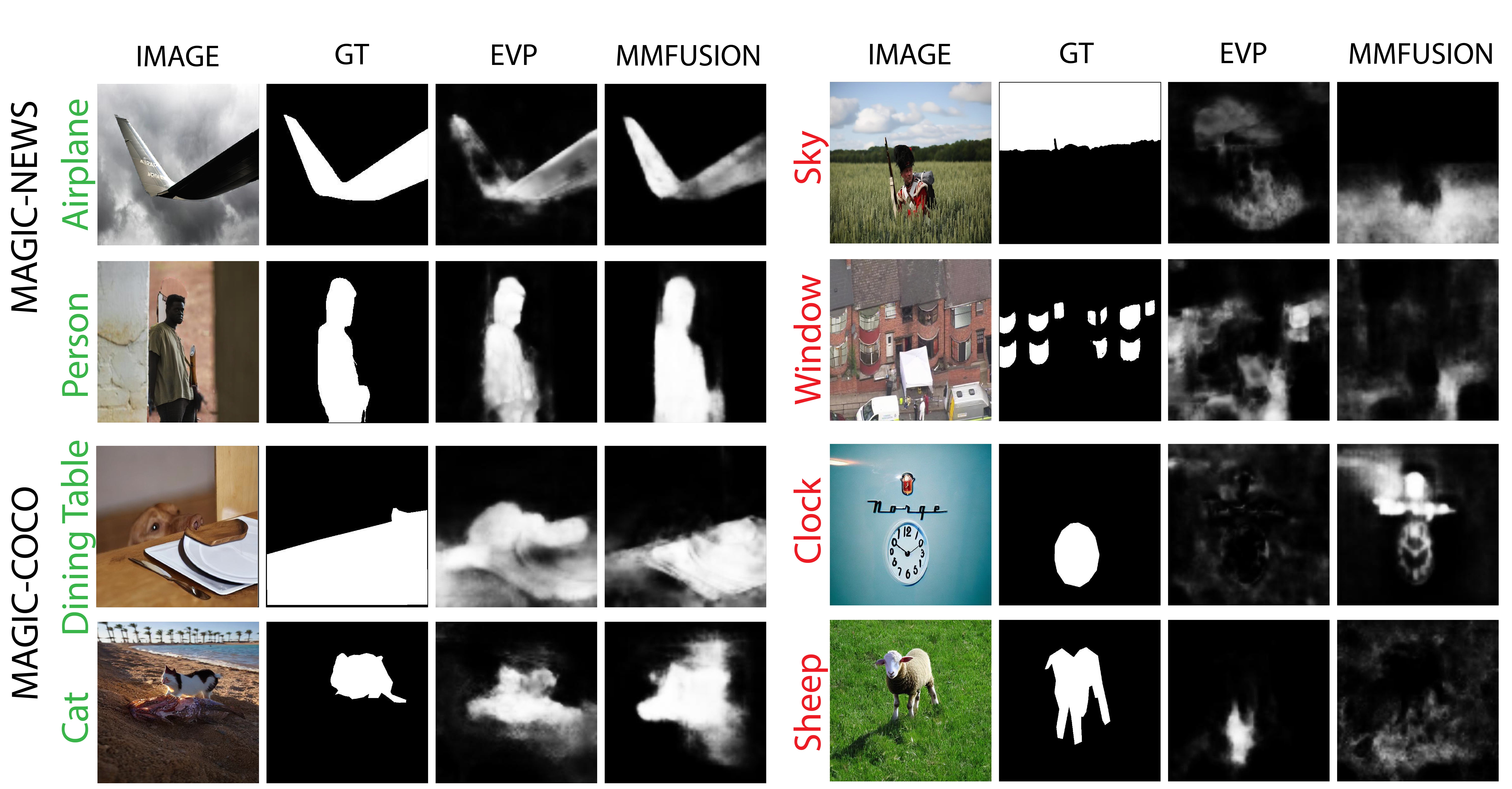} 
    \vspace{-2mm}
    \caption{Qualitative comparison of EVP and MMFusion across object categories from \DataShort{}-NEWS (top) and \DataShort{}-COCO (bottom). These results suggest that both models struggle with 
similar object types under distribution shifts.}
    \label{fig:qualitative_results}
\end{figure*}

An additional important question we explore is how manipulation quality affects these detectors.  Low quality manipulations may be easy for humans to detect, suggesting high quality manipulations are more important. However, we argue against this interpretation as many applications require a detector that can detect both low and high quality manipulations.  For example, when trying to detect misinformation by identifying whether or not an image has been manipulated, users would not trust a system if the detector could not recognize that an image has been manipulated when it is easy for them to identify.  Thus, we explore how human perception of manipulation quality by separating samples into "High" and "Low" quality manipulations using the results of our human study. Table~\ref{tab:survey_results_labelled_test} generally reports negligible impact between the two types of samples, with detectors often reporting performance with 2\% of each other (and quite often less than 0.5\%).  Notably, the ranking of methods did not change between the two manipulation types, \ie, this suggests that the manipulations types in our dataset are sufficient quality that detectors find that human perception has a negligible impact on their performance.  This also suggests that detectors may boost performance by trying to emulate human perception.

\subsection{Qualitative Analysis}
To further understand the behavior of our selected models, we conducted a qualitative analysis. We examined the manipulated object categories that appeared more than 500 times in each OOD dataset of \DataShort{}-NEWS and \DataShort{}-COCO and computed the average AUC for each object category. Our findings indicate that certain objects, such as "person," exhibited strong generalization, whereas other objects like "sky" or "sheep" showed performance degradation. Moreover, we compared results between EVP and MMFusion and observed that for many object categories, their performance remained similar. These insights suggest that certain objects can be challenging to detect across models; additional analysis is provided in Section~\ref{sec:object_cat_analysis_appendix} of the Appendix. 

%% file: sec/5_conclusion.tex
\section{Conclusion}

We introduced \DataLong{} (\DataShort{}), a large image manipulation dataset to study the robustness and generalization capabilities of image manipulation detectors. Our dataset features a range of image sources, manipulation types, and sizes. Notably, we have utilized state-of-the-art image manipulation techniques. Through extensive experiments, we found that while current manipulation localization methods perform well on in-distribution data, they struggle on out-of-distribution samples, underscoring the need for better generalization. 

In our future research we will focus on integrating this contextual information to add to detection models and enhancing the quality of manipulations to drive further advancements in the field.

\section*{Acknowledgments}
This material is based upon work supported, in part, by DARPA under agreement number HR00112020054. Any opinions, findings, and conclusions or recommendations expressed in this material are those of the author(s) and do not necessarily reflect the views of the
supporting agencies.

%% file: x_supplementary.tex
\section*{Appendix}

In this Appendix we include additional information about our experiments: 1) We conduct another analysis of our human eval results by looking at the High versus Low quality images. 2) We utilize the captions that are included with our \DataShort{}-News and \DataShort{}-COCO dataset to determine if their is any discrepancy between captions that are related or not to the manipulated object. 3) We include an example image of what a human annotator would have seen during the human evaluation. Additionally, we include an example of what the manually create Adobe Firefly manipulations looked like before and after and discuss the ethical considerations of our work.

\section{Ethical Considerations}
\label{sec:ethical_consi}

Our work focuses on benchmarking and advancing the methods for detecting manipulated images. We believe this is an important effort aimed at countering misinformation online, especially in the era of advanced generative models. Thus, this work is ethical by its nature. At the same time, since we are producing a dataset with manipulated images, there is some potential for misuse, \ie, it being used for training even more sophisticated falsification methods. However,  highlighting these issues should help inform experts when creating defenses.

\begin{table*}[t]
    \centering
      \setlength{\tabcolsep}{1.pt}
    \caption{Comparing AUC and F1 score of models trained only on \DataShort{}-News or \DataShort{}-COCO and tested on both to measure generalization across image sources and manipulation types this set contains Authentic Images. We bold the best numbers in the top half, whereas we underline the best method between EVP and combinations with DG methods in the bottom half.}
    \begin{tabular}{l cccc cccc cccc cccc}
    \toprule
    Trained on: & \multicolumn{4}{c}{\DataShort{}-News} & \multicolumn{4}{c}{\DataShort{}-COCO} & \multicolumn{4}{c}{\DataShort{}-COCO} & \multicolumn{4}{c}{\DataShort{}-News}\\
    \midrule
    Tested on: & \multicolumn{8}{c}{\DataShort{}-News}  & \multicolumn{8}{c}{\DataShort{}-COCO}\\
    \midrule
     & \multicolumn{2}{c}{MT-ID} & \multicolumn{2}{c}{MT-OOD} & \multicolumn{2}{c}{MT-ID} & \multicolumn{2}{c}{MT-OOD} & \multicolumn{2}{c}{MT-ID} & \multicolumn{2}{c}{MT-OOD} & \multicolumn{2}{c}{MT-ID} & \multicolumn{2}{c}{MT-OOD}\\
     \midrule
     & AUC & F1 & AUC & F1 & AUC & F1 & AUC & F1 & AUC & F1 & AUC & F1 & AUC & F1 & AUC & F1\\
     \midrule
        Upernet           &64.2&48.9&61.4&35.2&60.2&35.0&53.5&8.7&59.2&29.1&53.1&7.0&56.2&25.0&50.7&18.0\\
        DOLOS       &64.1&48.0&60.8&43.4&53.2&40.6&56.9&43.0&60.4&21.8&61.8&24.5&56.2&27.7&44.5&25.0\\
        PSCC-Net     &74.8&70.4&56.8&50.0&66.0&53.5&47.4&44.9&70.3&36.9&72.0&32.2&72.5&45.8&67.7&30.0\\
        MMFusion        &\textbf{83.0}&\textbf{81.4}&76.9&\textbf{82.6}&\textbf{76.9}&\textbf{70.8}&59.1&41.2&\textbf{80.9}&\textbf{64.9}&68.1&27.5&74.4&\textbf{51.2}&59.0&23.5\\
        HiFi         
        &53.9&14.0&50.2&1.5&57.2&39.6&51.7&39.6&63.7&14.0&56.5&35.1&51.6&6.2&49.8&0.4\\
        EVP          
        &80.9&76.6&\underline{\textbf{81.5}}&\underline{61.6}&68.3&54.2&\textbf{64.6}&\textbf{45.5}&76.8&49.1&\textbf{81.7}&\underline{\textbf{39.1}}&71.6&45.7&65.3&\underline{29.3}\\
        TruFor
        &51.8&40.6&51.2&41.7&50.7&41.0&48.6&40.8&50.3&21.3&48.5&21.2&54.7& 24.9&53.4&24.6\\
        SIDA
        &73.2&68.2&74.8&59.1&62.0& 41.6&61.6&33.3&66.5&43.5&63.5&33.0&67.9& 46.8&\textbf{73.9}&\textbf{46.2}\\ 
        
        \midrule
        EVP+SWAD
        &76.7&65.6&78.0&57.2&70.8&56.3&64.5&45.7&77.2&49.6&77.1&36.0&66.7&37.1&62.3&28.0\\
        MMFusion+SWAD
         &84.0&81.3&80.4&64.0&77.5&69.2&64.3&44.4&82.3&65.9&74.2&36.9&75.5&50.4&59.2&29.0\\
        EVP+Soup & 77.7&70.1&77.7&54.6&70.1&56.9&64.3&\underline{45.9}&77.8&49.3&76.1&34.5&68.9&39.2&63.4&28.3\\
        EVP+MIRO &\underline{82.0}&\underline{78.0}&76.8&53.9&73.0&59.2&63.3&44.9 &79.4&53.0&\underline{77.4}&32.9&\underline{75.2} &\underline{49.8} & 62.9&26.9\\
        MMFusion+MIRO
          &83.6&81.7&74.1&61.3&77.3&69.6&61.4&43.3&83.1&66.9&74.6&36.1&72.0& 48.9&47.5&27.0\\
        EVP+URM &80.0&70.4&73.9&51.1&\underline{74.9}&\underline{64.0}&\underline{64.9}&44.4&\underline{79.7}&\underline{56.6}&73.6&30.5&71.9&43.9&\underline{65.5}&25.6\\

       \bottomrule
    \end{tabular}
      \vspace{-4mm}
    \label{tab:vn_news_results_authentic}
\end{table*}

\section{Results with Authentic Images}
This can be seen in Table~\ref{tab:vn_news_results_authentic} which includes the results with authentic images when calculating the AUC. Because of the fact that in many cases the AUC for authentic images was around 50\% for the AUC we reported the model performance with authentic images in this table. We can see that in some cases for example with EVP when looking at columns 1 and 3 it appears that the AUC is higher for MT-ID and MT-OOD, however when we consider the fact the MT-ID contains authentic images, it helps explain why the performance appears lower. This is why we report Table~\ref{tab:vn_news_results} in the main paper which helps us better understand the impact in performance with only manipulations. One thing that is still highlighted in Table~\ref{tab:vn_news_results_authentic} is that the image source performance still drops as can be seen for MMFusion in columns 1 and 5 which highlights the impact of image source and showcases a real world setting which manipulated images can come from many different source.

\begin{table*}[t]

   \caption{Comparing the Mean and Standard Deviation (SD) of AUC for models trained using 3 different seeds for models trained on  \DataShort{}-News and \DataShort{}-COCO, respectively, and tested on both sets to measure generalization across image sources and manipulation types. Model used are EVP~\citep{evp}, MMFusion~\citep{mmFusion} EVP+MIRO~\citep{miro}.}
    \centering
      \setlength{\tabcolsep}{1.2pt}
    \begin{tabular}{l cccc cccc cccc cccc}
    \toprule
    Trained on: & \multicolumn{4}{c}{\DataShort{}-News} & \multicolumn{4}{c}{\DataShort{}-COCO} & \multicolumn{4}{c}{\DataShort{}-COCO} & \multicolumn{4}{c}{\DataShort{}-News}\\
    \midrule
    Tested on: & \multicolumn{8}{c}{\DataShort{}-News}  & \multicolumn{8}{c}{\DataShort{}-COCO}\\
    \midrule
     & \multicolumn{2}{c}{MT-ID} & \multicolumn{2}{c}{MT-OOD} & \multicolumn{2}{c}{MT-ID} & \multicolumn{2}{c}{MT-OOD} & \multicolumn{2}{c}{MT-ID} & \multicolumn{2}{c}{MT-OOD} & \multicolumn{2}{c}{MT-ID} & \multicolumn{2}{c}{MT-OOD}\\
     \midrule
     & Mean & SD & Mean & SD & Mean & SD & Mean & SD & Mean & SD & Mean & SD & Mean & SD & Mean & SD\\

     \midrule
        EVP            
        &81.1&0.54&79.3& 0.97&71.7&0.44&64.5&1.2&79.8&0.14&81.5&0.81&71.8&1.50&61.5&2.77\\
        MMFusion            
        &84.0&0.08&79.5&1.92&76.9&0.69&64.6&2.09&82.3&0.41&76.1&2.05&76.3&0.97&61.9&2.24\\
        
       \bottomrule
    \end{tabular}
     
    \label{tab:audits_multiple_experiments}
\end{table*}

\section{Semantic Saliency}

{\noindent\textbf{Manipulation Semantic Salience}}
Another aspect of our dataset is looking at the semantic saliency with respect to captions, namely if a manipulated object is mentioned in a caption describing a manipulated image. For this task we utilize the original captions used from Visual News and COCO.

\begin{table*}[t]
\caption{Comparing the AUC performance of models trained on the \DataShort{}-News and \DataShort{}-News respectively subset to generalize across image distributions and caption relevance. *Cap-Ref: manipulated object in caption *Not Ref: manipulated object not in caption. Columns 2-5 are tested on the \DataShort{}-News test set and columns 6-9 are tested on the \DataShort{}-COCO test set.}
\centering
    \begin{tabular}{lcccccccc}
    \toprule
    & \multicolumn{2}{c}{\DataShort{}-News} & \multicolumn{2}{c}{\DataShort{}-COCO}& \multicolumn{2}{c}{\DataShort{}-News} & \multicolumn{2}{c}{\DataShort{}-COCO}\\
     \midrule
        & Cap-Ref & Not Ref & Cap-Ref & Not Ref & Cap-Ref & Not Ref & Cap-Ref & Not Ref \\
     \midrule
        EVP      &\textbf{75.3}&\textbf{75.0}&60.7&57.1&\textbf{89.6}&\textbf{86.1}&\textbf{66.2}&\textbf{67.5}\\
        DOLOS    &72.6&71.8&\textbf{67.9}&\textbf{66.5}&64.1&63.1&49.8&50.5\\
        PSCC-Net &65.2&64.9&51.2&50.2&75.6&74.5&49.2&49.4\\
       \bottomrule
    \end{tabular}
    \label{tab:vn_relevance_captions}
\end{table*}

{\noindent\textbf{Generalizing across manipulation semantic salience}}
Table~{\ref{tab:vn_relevance_captions}} refers to the results obtained for related and unrelated captions with respect to the manipulated object being mentioned in the caption. We can see a similar trend to Table 2 in the main paper whereby for both related and unrelated captions EVP tends to be the best performing model for columns 2-3 and 6-9 as to be expected with DOLOS performing the best for columns 4-5. For EVP, columns 4-5 and 6-7 we can see that the *Cap-Ref were the only times it scored slightly higher than *Not Ref, we know for \DataShort{}-News there are more related captions hence the higher performance for columns 4-5 are expected. However for columns 6-7, being tested on the \DataShort{}-COCO test set showcases again how challenging the \DataShort{}-COCO subset can be, as we can see there are fewer related captions.  Hence, having a model that has a loosely related caption can possibly highlight challenging manipulations to detect even with the best performing manipulation detection models.

\begin{table*}[t]
\caption{Number of images per topic in our \DataShort{}-News subset.}
\label{tab:topic-count}
\centering
\begin{tabular}{ccccccc}
    \toprule
    International & Law and Crime & Arts & World & Science & Sport & Conflict\\
    \midrule
    36,684 & 35,273 & 35,867 & 35,489 & 33,541 & 24,022  & 8,188 \\
    \toprule
    Nature & Film & Music & Business & Politics & Disaster & \\
    \midrule
    24,157 & 19,694 & 17,374 & 16,155 & 15,080 & 10,382 & \\
    \bottomrule
\end{tabular}
\end{table*}

\begin{table}[t]
\centering
\caption{Number of images per news outlet in our \DataShort{}-News subset.}
\label{tab:news_source_counts}
\begin{tabular}{lc}
\toprule
Source & Number of Images \\
\midrule
The Guardian & 120,753 \\
BBC & 72,201 \\
USA Today & 62,682 \\
Washington Post & 56,401 \\
\bottomrule
\end{tabular}
\end{table}

\subsection{AUDITS-News Dataset Statistics}
\label{sec:audits_news_stats}

We provide additional statistics for the AUDITS-News subset derived from the 
VisualNews dataset. Table~\ref{tab:topic-count} shows the distribution of images 
across different news topics. The dataset spans a wide range of categories 
including international news, law and crime, arts, science, and politics, 
ensuring diversity in scene content and semantic context. This variety helps 
capture realistic image manipulation scenarios that arise in real-world 
journalistic settings.

Table~\ref{tab:news_source_counts} presents the distribution of images across news 
sources. The dataset includes images from four major outlets: The Guardian, BBC, 
USA Today, and The Washington Post. While the distribution is not perfectly 
balanced, all sources contribute a substantial number of images, providing 
diversity in capture styles, editorial practices, and visual content. This 
variation helps reduce bias toward a single publication and improves the 
robustness of evaluation across different news domains.

\begin{figure*}[t] 
  \centering
  \includegraphics[width=1.0\textwidth]{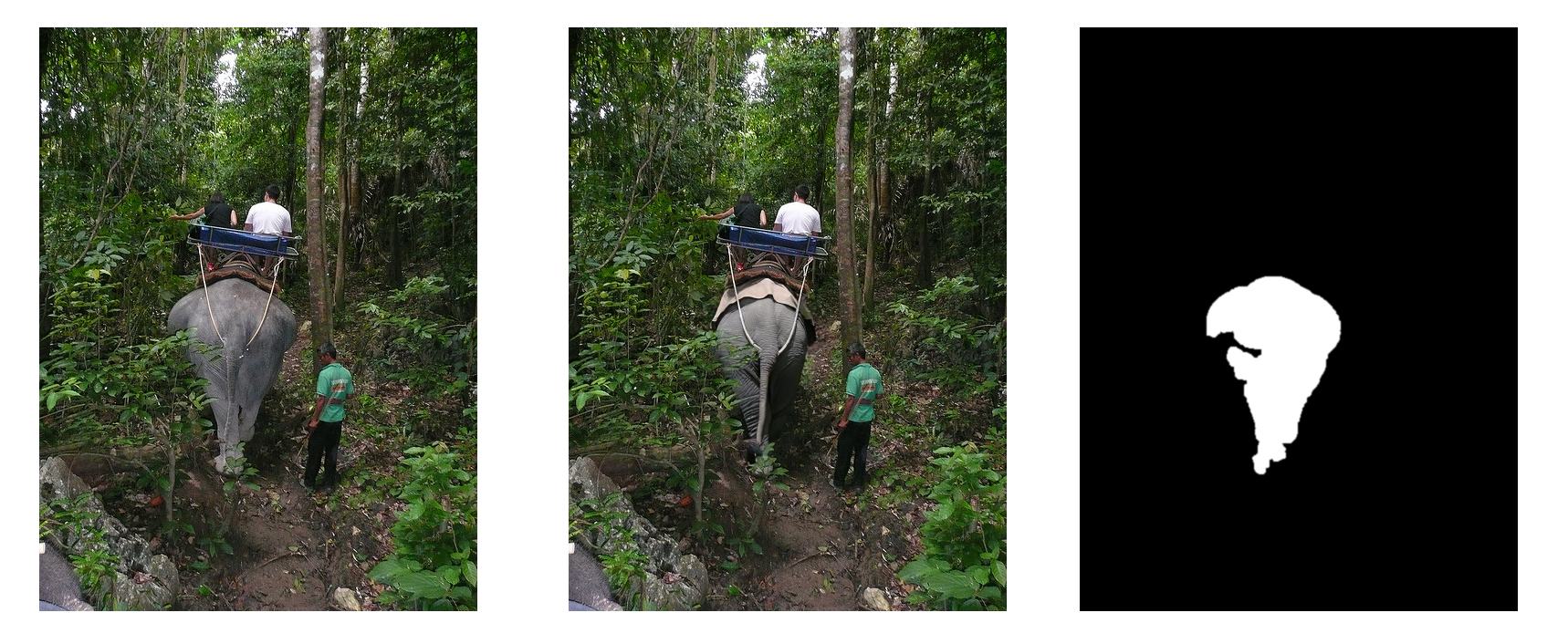} 
  \caption{An example of a Adobe Firefly inpainted image from COCO  described in Section~\ref{sec:replace_removal_manips} in the main paper. With the left most being the original and the middle being the Firefly Inpainted image.}
  \label{fig:adobe_firefly_image}
\end{figure*}

\subsection{ Traditional Image Manipulation Analysis}
We include a different experiment whereby we investigate the performance of the EVP~\citep{evp} model trained on data from either DEFACTO, \DataShort{} or both. We can see the performance on traditional image manipulation like Copymove and Splicing in Table~\ref{tab:traditional_performance_evp}. Based on these results we can see that for traditional manipulations training on either the DEFACTO dataset or the AUDITS+DEFACTO dataset provides much better results for both CASIA and DEFACTO, this is of course expected since our dataset is not composed of any taditional manipulations. However we see the drop in performance when trying to train using traditional manipulation and testing on diffusion based inpaintings as we can see in Table~\ref{tab:traditional_peformance_audits_evp}. We can see training on DEFACTO has a big drop in performance when testing on diffusion based inpaintings, which highlights the benefit of our dataset. Additionally we can see combining our \DataShort{} dataset with other traditional manipulation datasets like DEFACTO can provide comparable performance for both traditional and diffusion based manipulations as is seen in both Table~\ref{tab:traditional_performance_evp} and Table~\ref{tab:traditional_peformance_audits_evp}.

\begin{table*}[t]
    \centering
    \caption{Comparing the performance of the EVP~\citep{evp} model trained on data from either DEFACTO~\citep{DEFACTO}, \DataShort{}-COCO or both and testing on classic image manipulation datasets that contain Copymove (CM) and Splicing (SP) images, namely CASIAv1, CASIAv2~\citep{CASIAV1V2} and DEFACTO  }
    \begin{tabular}{l cccc cccc cccc}
    \toprule
    & \multicolumn{12}{c}{Tested on:} \\
    \midrule
    
    & \multicolumn{4}{c}{CASIAv1 }& \multicolumn{4}{c}{CASIAv2 }& \multicolumn{4}{c}{DEFACTO}\\
               &  \multicolumn{2}{c}{CM}   & \multicolumn{2}{c}{SP}                  & \multicolumn{2}{c}{CM}   & \multicolumn{2}{c}{SP} &                     \multicolumn{2}{c}{CM}   & \multicolumn{2}{c}{SP}         \\
               & AUC & F1 & AUC & F1 & AUC & F1 & AUC & F1 & AUC & F1 & AUC & F1 \\
    Trained on: &     &    &     &    &     &    &     &    &     &    &     &    \\   
    AUDITS+DEFACTO & \textbf{67.9} &20.3 & 86.9 &47.6 & 65.3 & 15.1 & 72.8 & 35.0 &  85.3 & 18.0 & 96.0& 93.7   \\
    DEFACTO & 63.5 & 18.1 & \textbf{87.0} & \textbf{51.3} & \textbf{67.8}& \textbf{16.4} & \textbf{76.0} & \textbf{39.5} & \textbf{88.7} & \textbf{22.9}  & \textbf{97.1} & \textbf{67.5}   \\
    AUDITS & 67.1 & \textbf{20.9}  & 82.2 & 40.3  & 56.8 & 12.8  & 54.3 & 25.3  & 53.1 & 4.9  & 65.6 & 9.5   \\
    \midrule
    \label{tab:traditional_performance_evp}
    \end{tabular}
\end{table*}

\begin{table*}[t]
    \centering
    \caption{Comparing the performance of the EVP~\citep{evp} model trained on data from either DEFACTO~\citep{DEFACTO}, \DataShort{}-COCO or both and testing on our \DataShort{} dataset to determine the diffusion based inpainting performance}

    \begin{tabular}{l cccc cccc}
    \toprule
    & \multicolumn{8}{c}{Tested on:} \\
    \midrule
    & \multicolumn{4}{c}{\DataShort{}-News}& \multicolumn{4}{c}{\DataShort{}-COCO}\\
               &  \multicolumn{2}{c}{MT-ID} & \multicolumn{2}{c}{MT-OOD} & \multicolumn{2}{c}{MT-ID} & \multicolumn{2}{c}{MT-OOD} \\
               & AUC & F1 & AUC & F1 & AUC & F1 & AUC & F1  \\
    Trained on: &     &    &     &    &     &    &     &    \\
    AUDITS+DEFACTO & 72.6 & 50.6 & \textbf{65.0} & 44.4  & \textbf{88.8} & \textbf{51.3}  & \textbf{82.1} & \textbf{41.8}  \\
    DEFACTO  & 55.4 & 29.3  & 55.4 & 29.5  & 68.7 & 28.2  & 72.7 & 32.7  \\
    AUDITS  & \textbf{74.4} & \textbf{54.2}  & 64.6 & \textbf{45.5}  & 85.8 & 49.1  & 81.7 & 39.1  \\
    \midrule
    \label{tab:traditional_peformance_audits_evp}
    \end{tabular}
    
\end{table*}

\section{Image-Level Detection}
To complement our localization experiments, we evaluate image-level manipulation detection on \DataShort{}. We evaluated the classification task using PSCC-Net~\citep{PSCCNet} and HiFi~\citep{Hifi}, as both models include a classification head. The results, summarized in Table~\ref{tab:classification_eval}, show that PSCC-Net outperforms HiFi in most cases despite their similar architecture.

Interestingly, PSCC-Net performs particularly well when trained and tested on \DataShort{}-COCO, likely due to its HRNet~\citep{HRNet} backbone, which is pre-trained on ImageNet~\citep{imagenet}. However, both models experience a significant drop in performance when tested on OOD images. This highlights the importance of our work in exposing these generalization challenges.

\begin{table*}[t]
\centering
\caption{Performance of PSCC-Net and HiFi on classification tasks.}
\label{tab:classification_eval}
\begin{tabular}{@{}llcccc@{}}
\toprule
Trained on &\DataShort{}-News&\DataShort{}-COCO&\DataShort{}-COCO&\DataShort{}-News \\ \midrule
Tested on & \multicolumn{2}{c}{\DataShort{}-News} & \multicolumn{2}{c}{\DataShort{}-COCO} \\ \cmidrule(lr){2-3} \cmidrule(lr){4-5}
& AUC/F1 & AUC/F1&AUC/F1 & AUC/F1\\ \midrule
PSCC-Net&69.0/64.1 &40.7/37.4&93.0/90.1&66.4/63.7\\
HiFi&54.7/54.7&50.7/34.9&56.0/55.8 &53.9/53.6 \\
\bottomrule
\end{tabular}
\end{table*}

\section{Qualitative Analysis of Object Categories}
\label{sec:object_cat_analysis_appendix}
\begin{figure*}[t]
    \centering
    \includegraphics[width=\textwidth]{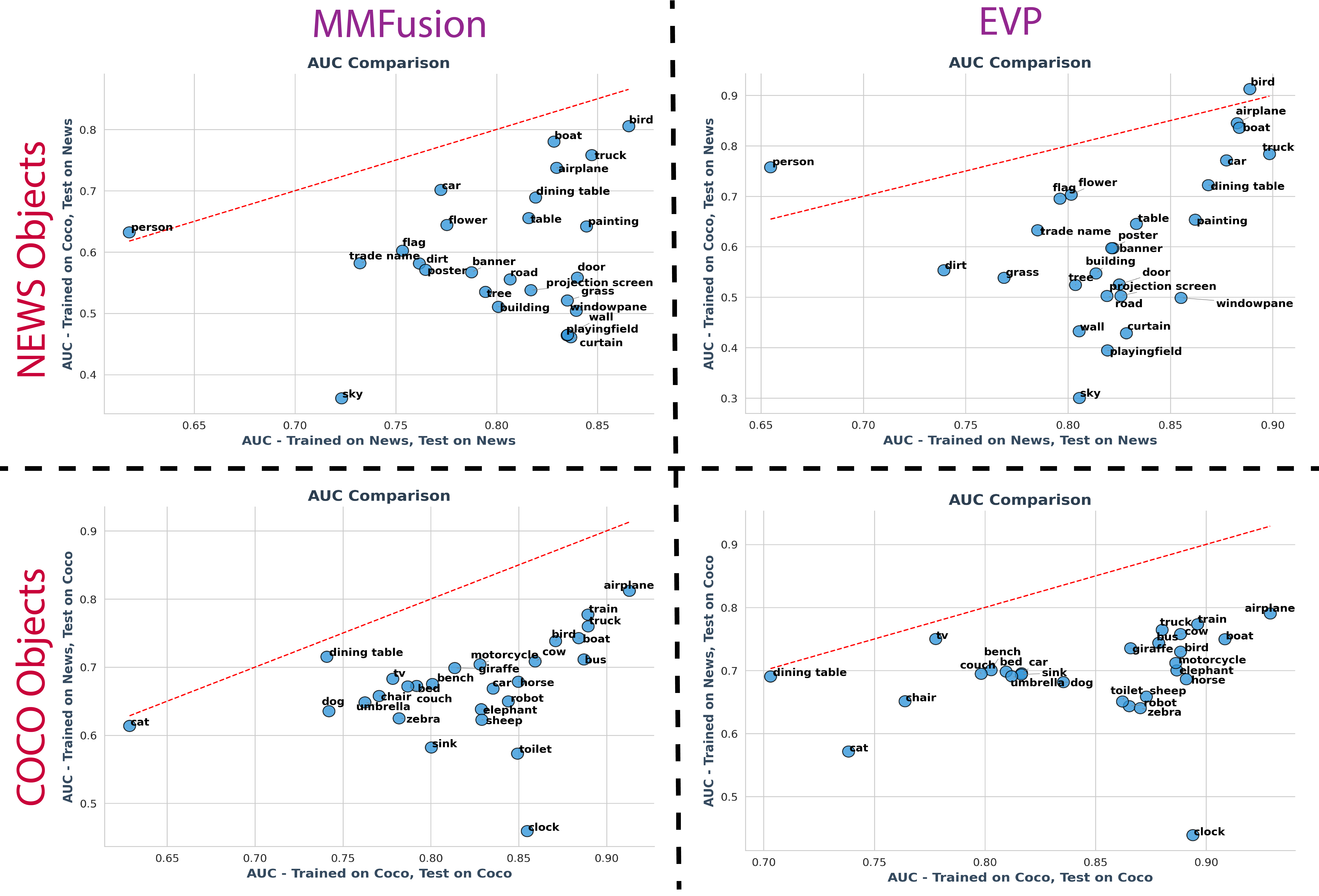} 
    \vspace{-4mm}
    \caption{Evaluation of mean AUC Object Categories using MMFusion and EVP}
    \label{fig:object_categories_models}
\end{figure*}

To further illustrate our qualitative findings, we plotted the average AUC for each selected object category across different training and testing setup for our two high performing models EVP and MMFusion. This can be seen in Figure~\ref{fig:object_categories_models}. The X-axis represents the AUC for models trained and tested on the same subset \DataShort{}-NEWS or \DataShort{}-COCO and Y-axis represents the AUC for models trained and tested on different subsets. From this analysis, we noted that objects such as "person" maintained high generalization performance across datasets, while certain object categories like "sky" exhibited a significant drop in performance. Additionally, we compared the performance of EVP and MMFusion across all object categories and found that their success and failure cases are same objects most times, suggesting that both models struggle with similar object types. 

\begin{figure*}
    \centering
    \includegraphics[width=1.0\textwidth]{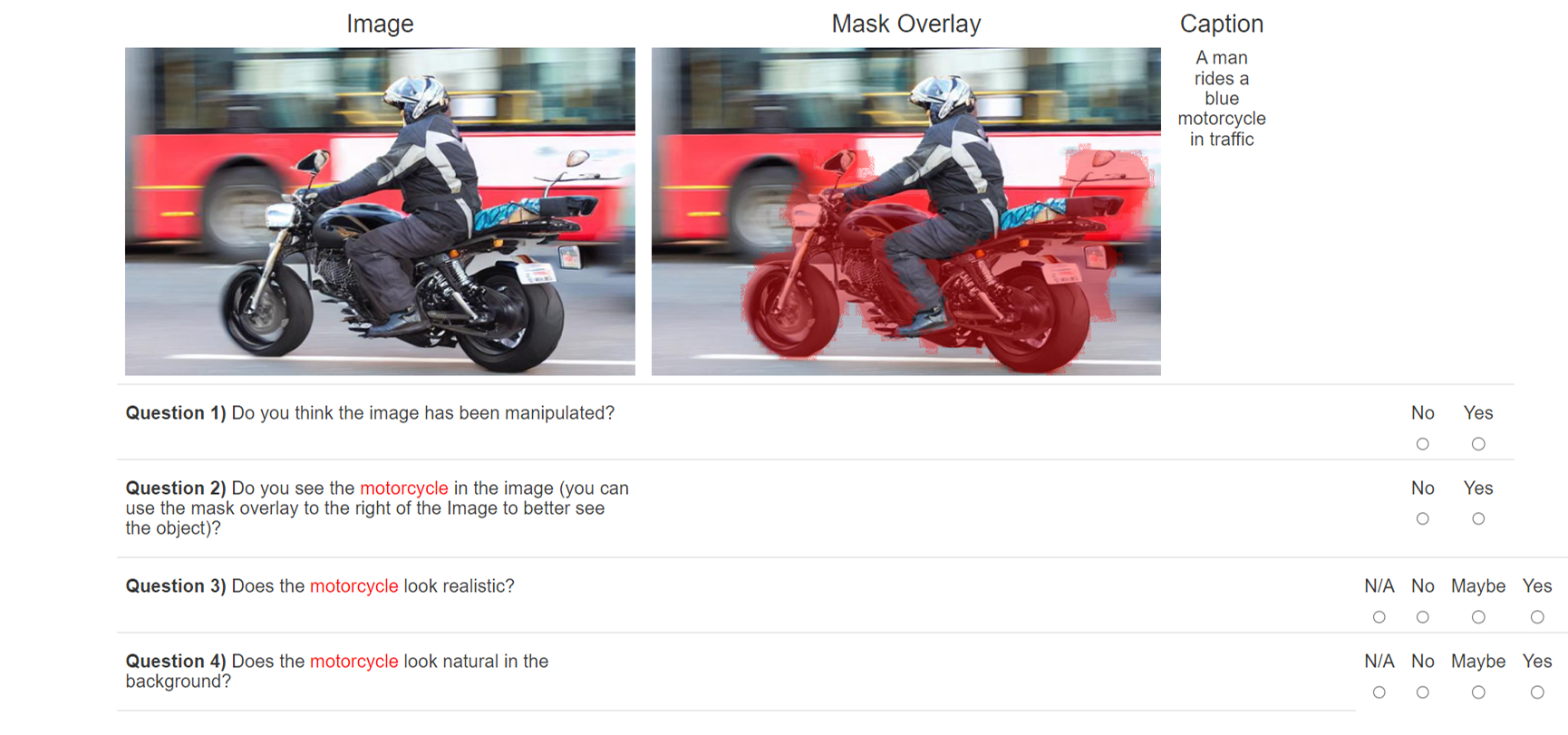} 
    \caption{An example of an image from our dataset which human evaluators were given to answer questions on.}
    \label{fig:human_eval_example}
\end{figure*}

\begin{figure*}[t] 
  \centering
  \includegraphics[width=0.8\textwidth]{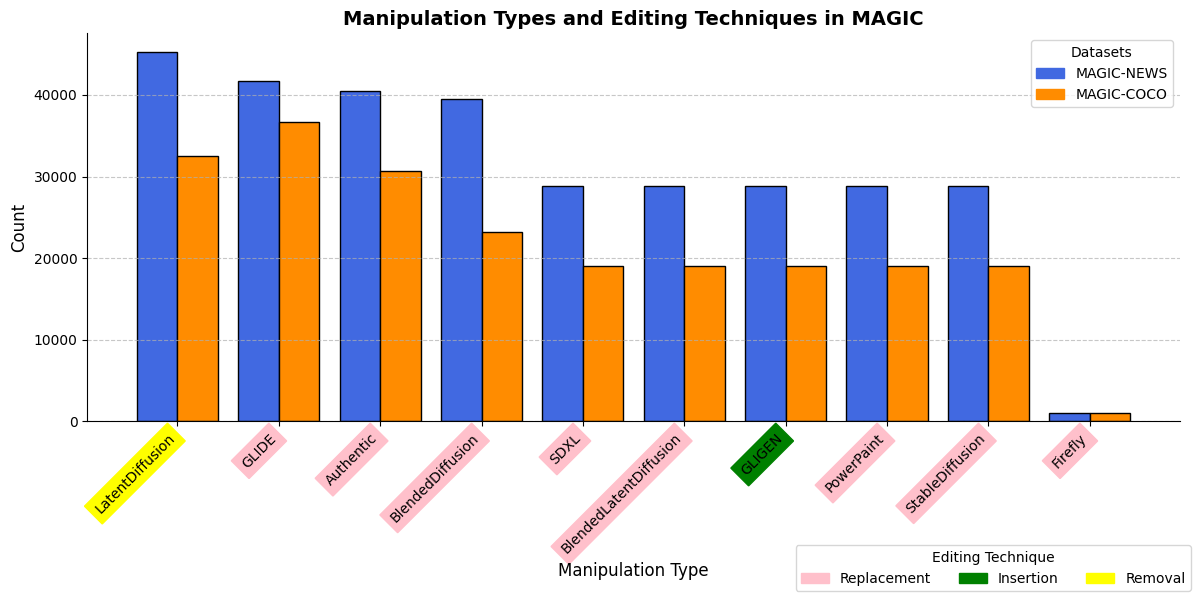} 
  \caption{Visualization of manipulation types across different editing techniques from in Figure\ 3 of the main paper.}
  \label{fig:editing_techniques_manip_size_}
\end{figure*}

\begin{figure*}[t] 
  \centering
  \includegraphics[width=0.8\textwidth]{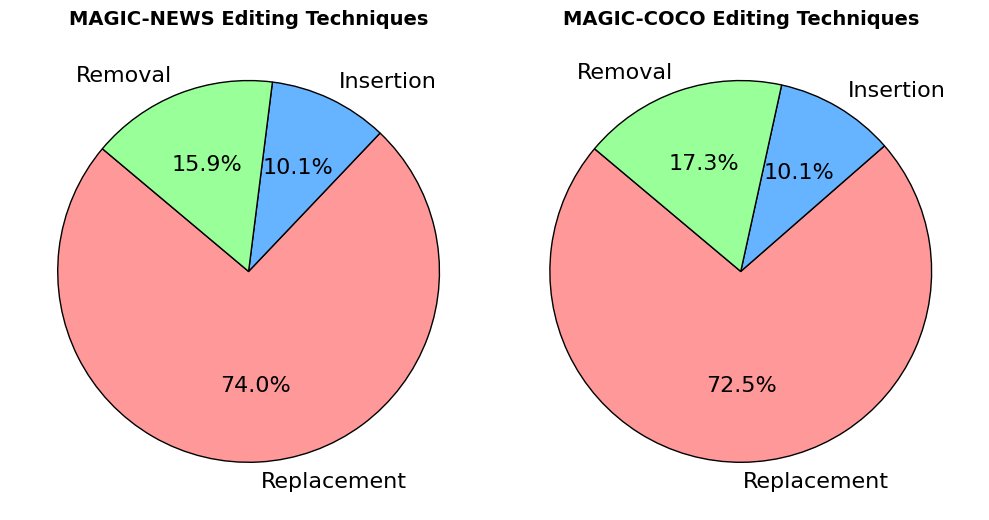} 
  \caption{Distribution of different editing techniques. }
  \label{fig:editing_techniques_manip_size}
\end{figure*}

\begin{figure*}[t] 
  \centering
  \includegraphics[width=0.6\textwidth]{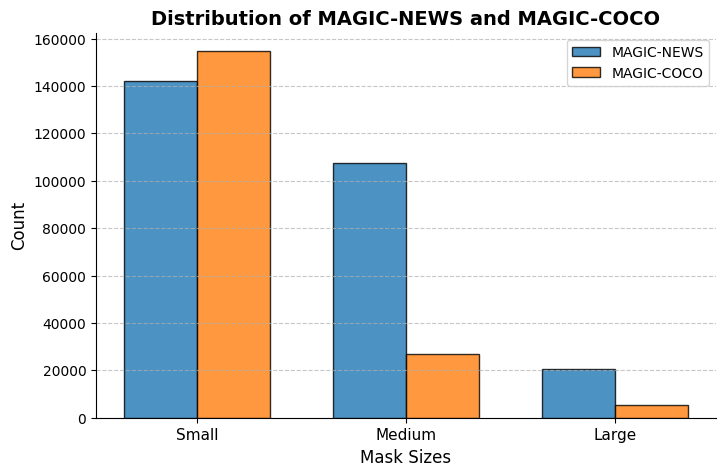} 
  \caption{Visualization of manipulation sizes across different image sources from Figure\ 3 of the main paper.}
  \label{fig:image_source_manip_size}
\end{figure*}

\subsection{Manipulation Size Distribution}
\label{sec:mask_size_distribution}

We analyze the distribution of manipulation sizes across the \DataShort{}-NEWS and \DataShort{}-COCO subsets. Figure~\ref{fig:image_source_manip_size} shows the number of manipulated regions categorized as small, medium, and large based on mask area. Both subsets are dominated by small manipulations, with fewer medium and large regions. We also observe differences in the relative proportions between \DataShort{}-NEWS and \DataShort{}-COCO. These variations arise from the underlying image content as well as the mask generation procedure. This visualization provides additional insight into the range of manipulation sizes present in \DataShort{}, which can influence the difficulty of manipulation localization.

\begin{table*}[t]
\centering
\caption{AUC/F1 Performance when evaluated on CasiaV1~\citep{CASIAV1V2} trained on \DataShort{}-News and \DataShort{}-COCO respectively}
\label{tab:casia_v1}
\begin{tabular}{@{}lccc@{}}
\toprule
Trained on \DataShort{}-News        & Copy-Move  & Splicing \\ \midrule
EVP     &   55.8/16.4 &64.3/31.6 \\
MMFusion & 	54.1/13.2 & 58.2/20.0 \\
PSCC-Net &	67.1/21.3 & 63.1/33.1 \\ \bottomrule
Trained on \DataShort{}-COCO        & Copy-Move  & Splicing \\ \midrule
EVP      &  67.1/20.9 & 82.2/40.3\\
MMFusion & 	49.2/13.2 & 53.3/19.5 \\
PSCC-Net &	56.8/16.5 & 67.5/29.3 \\ \bottomrule
\end{tabular}
\end{table*}

\begin{table*}[t]
\centering
\caption{AUC/F1 Performance when evaluated on CasiaV2~\citep{CASIAV1V2} trained on \DataShort{}-News and \DataShort{}-COCO respectively}
\label{tab:casia_v2}
\begin{tabular}{@{}lccc@{}}
\toprule
Trained on \DataShort{}-News        & Copy-Move  & Splicing \\ \midrule
EVP     &  	56.8/12.8  & 54.3/25.3	 \\
MMFusion & 		47.1/12.0 & 48.2/21.8 \\
PSCC-Net &		54.2/12.7 &  	50.2/27.1 \\ \bottomrule
Trained on \DataShort{}-COCO        & Copy-Move  & Splicing \\ \midrule
EVP      &  	56.8/12.8  &  54.3/25.3	\\
MMFusion & 	52.7/11.4  & 52.2/21.4  \\
PSCC-Net &		52.0/12.5  &  59.9/27.9 \\ \bottomrule
\end{tabular}
\end{table*}

\begin{table*}[t]
\centering
\caption{AUC/F1 Performance when evaluated on DOLOS~\citep{dolos} trained on \DataShort{}-News and \DataShort{}-COCO respectively}
\label{tab:dolos}
\begin{tabular}{@{}lccccc@{}}
\toprule
Trained on \DataShort{}-News        & LDM  & LAMA & Pluralistic & Repaint-P2 \\ \midrule
EVP     &  	61.7/57.4		   & 	50.0/47.6	 	& 58.7/52.6 & 59.4/53.5	 \\
MMFusion & 	42.5/57.3		  &  42.6/53.7	 & 43.4/55.4 & 42.8/55.2\\
PSCC-Net &	49.9/0.01		  &  54.5/52.8	  &48.5/0.09  & 49.0/13.1 \\ \bottomrule
Trained on \DataShort{}-COCO        & LDM  & LAMA & Pluralistic & Repaint-P2 \\ \midrule
EVP      &  	41.8/41.0		   & 44.9/42.1	  	& 44.1/42.1	 & 46.0/43.4 \\
MMFusion & 	 	43.3/52.6  &  42.6/48.9      & 43.4/52.1 & 43.1/51.4\\
PSCC-Net &	50.3/0.04	   &   54.2/53.5  &50.5/0.06	 & 50.6/0.05\\ \bottomrule
\end{tabular}
\end{table*}

\begin{table*}[t]
\centering
\caption{AUC/F1 Performance when evaluated on TGIF~\citep{TGIF} trained on \DataShort{}-News and \DataShort{}-COCO respectively}
\label{tab:tgif}
\begin{tabular}{@{}lccccc@{}}
\toprule
Trained on \DataShort{}-News        & PS-SP  & 	SD2-SP & SD2-FR & SDXL-FR \\ \midrule
EVP     & 65.8/14.0	 & 	66.1/14.3 &  66.6/27.0	&		67.3/16.4 \\
MMFusion & 57.9/13.8 & 57.1/13.5 & 53.5/25.1	 & 	53.3/12.8\\
PSCC-Net & 77.2/25.5 & 62.3/11.4	 & 50.0/0.01 & 49.9/0.02 \\ \bottomrule
Trained on \DataShort{}-COCO        & PS-SP  & 	SD2-SP & SD2-FR & SDXL-FR \\ \midrule
EVP      & 	61.8/13.5	 & 62.7/14.2	 & 65.5/26.5 &  71.3/17.4	\\
MMFusion & 44.8/12.6	 & 46.7/12.9 & 47.9/24.3 &	47.5/12.2	\\
PSCC-Net & 65.6/14.6 & 58.1/13.2	 & 51.4/0.02	 & 49.6/0.02\\ \bottomrule
\end{tabular}
\end{table*}

\begin{table*}[t]
\centering
\caption{AUC/F1 Performance when evaluated on IMD2020~\citep{imd2020} trained on \DataShort{}-News and \DataShort{}-COCO respectively}
\label{tab:imd2020}
\begin{tabular}{@{}lcc@{}}
\toprule
Trained on \DataShort{}-News        & Manipulated \\ \midrule
EVP     &  	53.5/51.1	 	 \\
MMFusion & 		50.8/55.1	  \\
PSCC-Net &		49.1/40.5	  \\ \bottomrule
Trained on \DataShort{}-COCO        & Manipulated \\ \midrule
EVP      &  42.4/36.6	 \\
MMFusion & 		53.2/53.9  \\
PSCC-Net &		41.1/46.9	 \\ \bottomrule
\end{tabular}
\end{table*}

\begin{table*}[t]
\centering
\caption{AUC/F1 Performance when evaluated on Autosplicing~\citep{autosplicing} trained on \DataShort{}-News and \DataShort{}-COCO respectively}
\label{tab:autosplicing}
\begin{tabular}{@{}lccc@{}}
\toprule
Trained on \DataShort{}-News        & JPEG100  & JPEG90 \\ \midrule
EVP     & 81.3/66.1 & 77.4/61.3 \\
MMFusion & 50.3/50.6 & 48.8/50.3 \\
PSCC-Net & 63.0/25.2 	& 81.6/65.5 \\ \bottomrule
Trained on \DataShort{}-COCO        & Copy-Move  & Splicing \\ \midrule
EVP      & 71.2/58.3 	&	82.7/63.5\\
MMFusion & 56.9/48.7 	& 58.5/49.0\\
PSCC-Net & 43.2/38.7 	& 75.8/61.5 \\ \bottomrule
\end{tabular}
\end{table*}

\section{Different Class Replacements in \DataShort{}-COCO and \DataShort{}-News}
We conducted a small-scale experiment to assess the impact of image class changes on model performance. Using 100 images from \DataShort{}-News, we created images where new objects replaced old ones (\eg replacing a car with a bike) while preserving semantic relevance. We performed the same experiment with \DataShort{}-COCO. Additionally, we included experiments with replacing objects with the same object type (\eg replacing a bike with another bike) as done in prior experiments.

The results, presented in Table~\ref{tab:news_replacement} and Table~\ref{tab:coco_replacement}, show that for both EVP and DOLOS, the performance differences between replacing objects with a new type versus the same type are minimal.

\begin{table*}[t]
\centering
\caption{Performance of EVP and DOLOS on \DataShort{}-News with object replacements.}
\label{tab:news_replacement}
\begin{tabular}{@{}lllll@{}}
\toprule
Trained on &\DataShort{}-News&\DataShort{}-News&\DataShort{}-COCO&\DataShort{}-COCO \\ \midrule
Tested on&\DataShort{}-News&\DataShort{}-News&\DataShort{}-News &\DataShort{}-News \\ \midrule
&New-Object&Replaced-Object&New-Object&Replaced-Object \\ \midrule
&AUC/F1/IoU &AUC/F1/IoU &AUC/F1/IoU &AUC/F1/IoU \\ \midrule
EVP&59.1/56.2/1.60&58.0/55.8/1.85&60.6/51.1/10.0&61.9/50.6/10.8 \\
DOLOS&54.1/54.6/5.72&55.0/54.9/6.20&51.5/58.3/2.66&47.7/58.4/2.15 \\
\bottomrule
\end{tabular}
\end{table*}

\begin{table*}[t]
\centering
\caption{The data quality survey outcomes, split by the inpainting type. We report majority vote for each question. Q1) \textit{``Do you think this image is manipulated?''}, Q2) \textit{``Do you see the \textbf{object} in the image (you can use the mask overlay to the right of the Image to better see the object)?''}; Q3) \textit{``Does the \textbf{object} look realistic?''}, and Q4) \textit{``Does the \textbf{object} look natural in the background?''}}
    \begin{tabular}{lccccc|cccc}
    \toprule
    & \multicolumn{5}{c|}{\DataShort{}-News} & \multicolumn{4}{c}{\DataShort{}-COCO}\\
    \midrule
    &GLIGEN&Blended&GLIDE&Latent&Stable&Blended&GLIDE&Latent&Stable\\
    \midrule
        Q1$\downarrow$&84.0&84.0&80.4&81.6&79.2&81.2&80.6&84.2&79.4\\
    Q2$\uparrow$&72.6&81.4& 75.4&72.0 & 82.0&81.2&80.6&84.2&79.4\\
    Q3$\uparrow$&48.5&51.0&59.5&54.6&54.1&57.1&60.8&54.8&61.2\\
    Q4$\uparrow$&56.1&54.2&63.1&59.5&65.2 &57.7&61.4&55.8&62.2\\
    \bottomrule
    \end{tabular}
   \label{tab:human_eval_survey_results}
\vspace{-4mm}  
\end{table*}

\section{Human Evaluation Results}

\label{sec:human_results_section}

When looking at Table~{\ref{tab:human_eval_survey_results}} we can see that for Q3, GLIGEN Splicing, Blended and Stable diffusion for News tended to be the worst performing model with a number of people realizing that those images were manipulated. On average for Q3 it appears that \DataShort{}-COCO tended to perform better with their manipulations for instance with GLIDE and Stable Diffusion being the top performing models. Q1 reveals that most persons did realize the image were manipulated which is to be expected as diffusion based inpaintings are still not perfect on average.

\begin{table*}[t]
    
    \centering
        \setlength{\tabcolsep}{5pt}
            \caption{The analysis of the model performance (AUC) based on our data quality survey outcomes.  
    We train our models on respective subsets and test them in domain on 1,121 images for \DataShort{}-News and 1,152 for \DataShort{}-COCO from the total 3,412 images from the Human Evaluation(see main text for discussion).
    }
    \begin{tabular}{lccc|ccc}
    \toprule
    Tested on: & \multicolumn{3}{c|}{\DataShort{}-News }& \multicolumn{3}{c}{\DataShort{}-COCO }\\
    \midrule
    Model & Blended & GLIDE & Latent& Blended & GLIDE & Latent\\ 
    \midrule
        EVP      &90.5&88.2&88.6&79.2&90.9&\textbf{86.1}\\
        DOLOS    &95.9&89.5&\textbf{93.4}&69.3&70.3&61.6\\
        PSCC-Net &\textbf{97.3}&\textbf{94.8}&85.0&\textbf{95.7}&\textbf{92.7}&68.5\\
       \bottomrule
    \end{tabular}

    \label{tab:survey_breakdown_results}
\end{table*}

In Section~\ref{sec:dataset_quality_survey} of our main paper we explained how we selected 3,412 images \footnote{The workers received \$0.1 for completing the survey.}. Next, we determine which images belong to the \DataShort{}-News and \DataShort{}-COCO ID test sets, and obtain 1,121 images for \DataShort{}-News and 1,152 for \DataShort{}-COCO. Table~\ref{tab:survey_breakdown_results} highlights the performance of the models trained in-domain and tested on these images. 
When comparing these results with those from {
Table~\ref{tab:human_eval_survey_results}} we see that even thought EVP and DOLOS tended to be the better performing models, when tested on \DataShort{}-COCO. They performed significantly worse than PSCC-Net for Blended Diffusion which could be explained by difficulty of of \DataShort{}-COCO in general. On average we see that models performed worse on \DataShort{}-COCO in Table~\ref{tab:survey_breakdown_results} versus \DataShort{}-News and this follows with what we see in {
Table~\ref{tab:human_eval_survey_results}} for Q3, as persons agreed that the manipulations from \DataShort{}-COCO were better on average. Hence we have shown that \DataShort{}-COCO is a more challenging dataset than \DataShort{}-News for diffusion based inpainting, and, therefore is a promising dataset for manipulation detection models to be tested on.

\begin{table*}[t]
\centering
\caption{Performance of EVP and DOLOS on \DataShort{}-COCO with object replacements.}
\label{tab:coco_replacement}
\begin{tabular}{@{}lllll@{}}
\toprule
Trained on &\DataShort{}-News&\DataShort{}-News&\DataShort{}-COCO&\DataShort{}-COCO \\ \midrule
Tested on&\DataShort{}-COCO&\DataShort{}-COCO&\DataShort{}-COCO &\DataShort{}-COCO \\ \midrule
&New-Object&Replaced-Object&New-Object&Replaced-Object \\ \midrule
&AUC/F1/IoU &AUC/F1/IoU &AUC/F1/IoU &AUC/F1/IoU \\ \midrule
EVP&87.0/53.1/35.0&86.9/53.2/34.8&60.8/26.0/1.81&61.8/25.9/0.77\\
DOLOS&67.2/25.0/2.93&66.8/26.1/2.85&44.8/23.7/1.62&46.1/23.9/1.21\\
\bottomrule
\end{tabular}
\end{table*}

\begin{table*}[t]
\centering
\caption{AUC Performance of EVP, MMFusion and PSCC-Net on recent diffusion based inpainting models when trained on \DataShort{}-News.}
\label{tab:recent_models_vs_news}
\begin{tabular}{@{}ll cccc cccc @{}}
\toprule
Trained on &\multicolumn{8}{c}{\DataShort{}-News} \\ \midrule
Tested on & \multicolumn{4}{c}{\DataShort{}-News} & \multicolumn{4}{c}{\DataShort{}-COCO} \\ \cmidrule(lr){2-5} \cmidrule(lr){6-10}
& Flux-Inpt & HD-Inpt & Firefly & SDXL & Flux-Inpaint & HD-Inpaint & Firefly & SDXL &\\ \midrule
EVP & 83.1 & 54.1 & 75.3 & 89.2 & 64.9 & 53.3 & 62.4 & 67.9\\
MMFusion & 47.6 & 52.4 & 51.2 & 80.3 & 57.3 & 51.9 & 51.9 & 57.4\\
PSCC-Net & 49.9 & 49.9 & 77.5 & 37.8 & 50.6 & 50.0 & 57.4 & 64.0 \\
\bottomrule
\end{tabular}
\end{table*}

\begin{table*}[t]
\centering
\caption{AUC Performance of EVP, MMFusion and PSCC-Net on recent diffusion based inpainting models when trained on \DataShort{}-COCO.}
\label{tab:recent_models_mscoco}
\begin{tabular}{@{}ll cccc cccc @{}}
\toprule
Trained on &\multicolumn{8}{c}{\DataShort{}-COCO} \\ \midrule
Tested on & \multicolumn{4}{c}{\DataShort{}-News} & \multicolumn{4}{c}{\DataShort{}-COCO} \\ \cmidrule(lr){2-5} \cmidrule(lr){6-10}
& Flux-Inpt & HD-Inpt & Firefly & SDXL & Flux-Inpaint & HD-Inpaint & Firefly & SDXL &\\ \midrule
    
EVP &65.6&47.1&63.7&68.3&54.9&41.4&50.3&77.5\\
MMFusion &46.3&47.3&47.5&59.0&52.9&55.4&47.2&60.4\\
PSCC-Net &48.3&49.8&44.2&39.9&49.3&49.4&59.2&71.0\\
\bottomrule
\end{tabular}
\end{table*}